%% file: eccv_paper.tex
\documentclass[runningheads]{llncs}

\usepackage[table, dvipsnames]{xcolor}
\usepackage[mobile]{eccv}

\usepackage{eccvabbrv}

\usepackage{graphicx}
\usepackage{booktabs}
\usepackage{comment}
\usepackage{lipsum} % Provides dummy text
\include{preamble}

\usepackage{wrapfig}
\usepackage{pifont}

\definecolor{cvprblue}{rgb}{0.21,0.49,0.74}
\definecolor{lblue}{rgb}{0.9,0.95,1}
\definecolor{lpurple}{rgb}{0.35,0.25,0.55}
\definecolor{lgreen}{rgb}{0.95,1,0.95}
\definecolor{sblue}{rgb}{0,0.45,1}

\usepackage[accsupp]{axessibility}  % Improves PDF readability for those with disabilities.

\usepackage[pagebackref,breaklinks,colorlinks,citecolor=eccvblue]{hyperref}
\usepackage{orcidlink}

\hypersetup{
     pdftitle={InstructIR: High-Quality Image Restoration Following Human Instructions},
     pdfsubject={Computer Vision, Image Restoration},
     pdfauthor={Marcos V. Conde, Gregor Geigle, Radu Timofte}
}

\begin{document}

\title{InstructIR: High-Quality Image Restoration Following Human Instructions}

\titlerunning{InstructIR}

\author{Marcos V. Conde\inst{1,2}\orcidlink{0000-0002-5823-4964} \and
Gregor Geigle\inst{1}%\orcidlink{1111-2222-3333-4444} 
\and
Radu Timofte\inst{1}\orcidlink{0000-0002-1478-0402}}

\authorrunning{Conde, Marcos et al.}

\institute{
Computer Vision Lab, CAIDAS \& IFI, University of Würzburg \and
Visual Computing Group, FTG, Sony PlayStation \\
\url{https://github.com/mv-lab/InstructIR} (500 \ding{80})
}

\maketitle

%%%%%%%%%%%%%%%%%%%%%%%%%%%%%%%%%%%%%%%%%%%%%%%%%%%%%%

\vspace{-2.5mm}
\begin{figure}
    \centering
    \includegraphics[width=\linewidth]{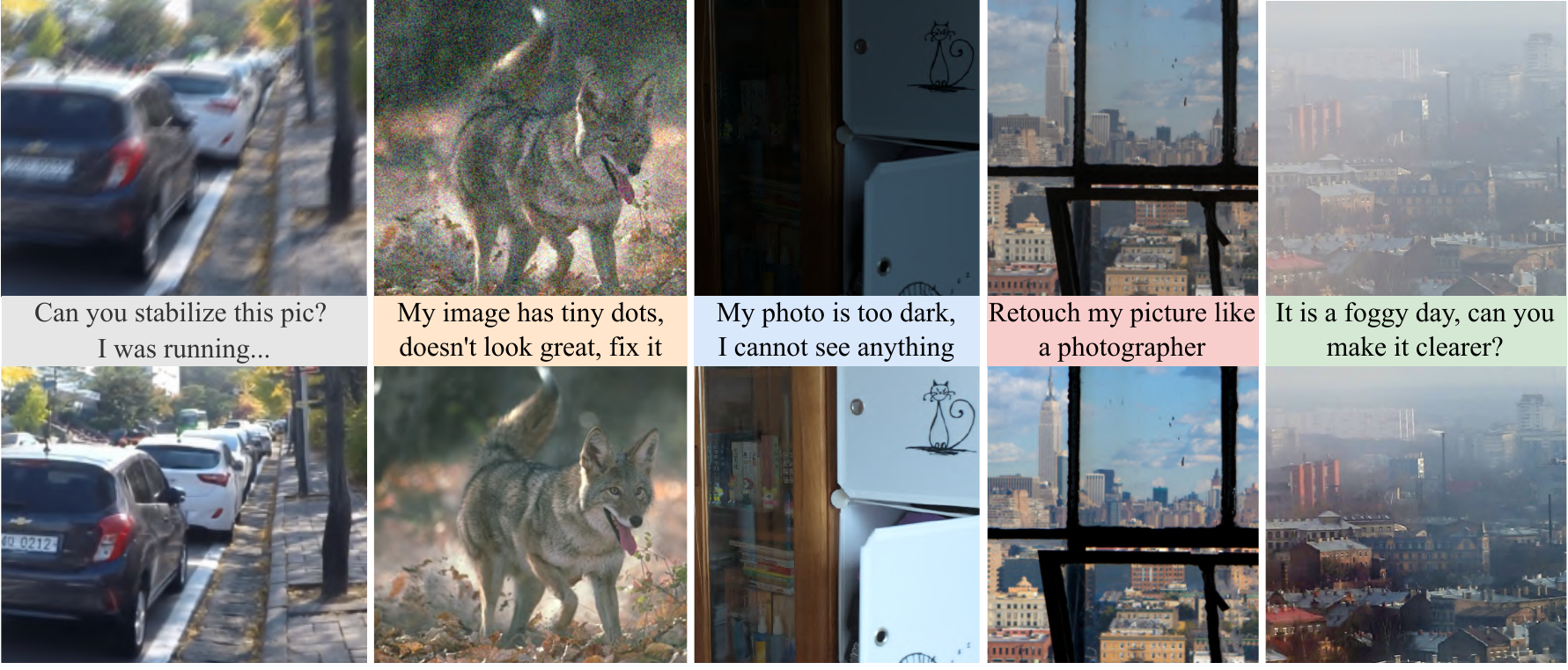}%
    \vspace{-2.3mm}
    \captionof{figure}{
    Given an {\bf image} and a {\bf prompt} for how to improve that image, our \emph{all-in-one} restoration model corrects the image considering the human instruction. \we, can tackle various types and levels of degradation, and it is able to generalize in some \emph{real-world} scenarios (last three images, from left to right).}
    \label{fig:teaser}
    \vspace{-7.5mm}
\end{figure}

\begin{abstract}
Image restoration is a fundamental problem that involves recovering a high-quality clean image from its degraded observation. All-In-One image restoration models can effectively restore images from various types and levels of degradation using degradation-specific information as prompts to guide the restoration model. In this work, we present the first approach that uses human-written instructions to guide the image restoration model. Given natural language prompts, our model can recover high-quality images from their degraded counterparts, considering multiple degradation types. Our method, InstructIR, achieves state-of-the-art results on several restoration tasks including image denoising, deraining, deblurring, dehazing, and (low-light) image enhancement. InstructIR improves +1dB over previous all-in-one restoration methods. Moreover, our dataset and results represent a novel benchmark for new research on text-guided image restoration and enhancement.
\end{abstract}

%%%%%%%%%%%%%%%%%%%%%%%%%%%%%%%%%%%%%%%%%%%%%%%%%%%%%%

\section{Introduction}

Images often contain unpleasant effects such as noise, motion blur, haze, and low dynamic range. Such effects are commonly known in low-level computer vision as \emph{degradations}. These can result from camera limitations or challenging environmental conditions \eg low light. 

Image restoration aims to recover a high-quality image from its degraded counterpart. This is a complex inverse problem since multiple different solutions can exist for restoring any given image~\cite{elad1997restoration, nguyen2001efficient, zhang2017beyond, zhang2017learning, dong2020multi, liang2021swinir}. 

Some methods focus on specific degradations, for instance reducing noise (denoising)~\cite{zhang2017beyond,zhang2017learning, ren2021adaptivedeamnet}, removing blur (deblurring)~\cite{nah2022clean,zhang2020deblurring}, or clearing haze (dehazing)~\cite{ren2020single,dong2020multi}. Such methods are effective for their specific task, yet they do not generalize well to other types of degradation. Other approaches use a general neural network for diverse tasks~\cite{tu2022maxim, restormer, wang2021uformer, chen2022simple}, yet training the neural network for each specific task independently. Since using a separate model for each possible degradation is resource-intensive, recent approaches propose \emph{All-in-One} restoration models~\cite{Li_2022_CVPR, potlapalli2023promptir, park2023all, zhang2023ingredient}. These approaches use a single deep blind restoration model considering multiple degradation types and levels. Contemporary works such as PromptIR~\cite{potlapalli2023promptir} or ProRes~\cite{ma2023prores} utilize a unified model for blind image restoration using learned guidance vectors, also known as ``prompt \emph{embeddings}", in contrast to raw user prompts in text form, which we use in this work.
%also known as multi-dimensional ``prompts". However, do not confuse these with real text prompts.

In parallel, recent works such as InstructPix2Pix~\cite{brooks2023instructpix2pix} show the potential of using text prompts to guide image generation and editing models. However, this method (or recent alternatives) do not tackle inverse problems. Inspired by these works, we argue that text guidance can help to guide blind restoration models better than the image-based degradation classification used in previous works~\cite{Li_2022_CVPR, zhang2023ingredient, park2023all}. Users generally have an idea about what has to be fixed (though they might lack domain-specific vocabulary) so we can use this information to guide the model.

\vspace{-2mm}
\paragraph{Contributions} We propose the first approach that utilizes real human-written instructions to solve multi-task image restoration.
Our comprehensive experiments demonstrate the potential of using text guidance for image restoration and enhancement by achieving \sota performance on various image restoration tasks, including image denoising, deraining, deblurring, dehazing, and low-light image enhancement.
Our model, \we, is able to generalize to restoring images using complex human-written instructions. Moreover, our single \emph{all-in-one} model covers more tasks than many previous works. We show diverse restoration samples of our method in Figure~\ref{fig:teaser}.

%%%%%%%%%%%%%%%%%%%%%%%%%%%%%%%%%%%%%%%%%%%%%%%%%%%%%%
\section{Related Work}
\label{sec:relwork}

\paragraph{Image Restoration.} Recent deep learning methods~\cite{dong2020multi,ren2021adaptivedeamnet,nah2022clean,liang2021swinir, restormer, tu2022maxim} have shown consistently better results compared to traditional techniques for blind image restoration~\cite{he2010single,dong2011image,timofte2013anchored,kim2010single,michaeli2013nonparametric,kopf2008deep}. The proposed neural networks are based on convolutional neural networks (CNNs) and Transformers~\cite{vaswani2017attention} (or related attention mechanisms). 
We focus on general-purpose restoration models~\cite{liang2021swinir, restormer, wang2021uformer, chen2022simple}. For example, SwinIR~\cite{liang2021swinir}, MAXIM~\cite{tu2022maxim} and Uformer~\cite{wang2021uformer}. These models can be trained -independently- for diverse tasks such as denoising, deraining or deblurring. Their ability to capture local and global feature interactions, and enhance them, allows the models to achieve great performance consistently across different tasks. For instance, Restormer~\cite{restormer} uses non-local blocks~\cite{wang2018non} to capture complex features across the image.

NAFNet~\cite{chen2022simple} is an efficient alternative to complex transformer-based methods. The model uses simplified channel attention, and gating as an alternative to non-linear activations. The building block (NAFBlock) follows a simple meta-former~\cite{yu2022metaformer} architecture with efficient inverted residual blocks~\cite{howard2019searching}. In this work, we build our \we model using NAFNet as backbone, due to its efficient and simple design, and high performance in several restoration tasks.

\vspace{-3mm}
\paragraph{All-in-One Image Restoration.}
Single degradation (or single task) restoration methods are well-studied, however, their real-world applications are limited due to the required resources \ie allocating different models, and selecting the adequate model on demand. Moreover, images rarely present a single degradation, for instance, noise and blur are almost ubiquitous in any image capture.

All-in-One (also known as multi-degradation or multi-task) image restoration is emerging as a new research field in low-level computer vision~\cite{Li_2022_CVPR, potlapalli2023promptir, park2023all, zhang2023all, zhang2023allamirnet, ma2023prores, yao2023neuraldegall, valanarasu2022transweather}. These approaches use a single deep blind restoration model to tackle different degradation types and levels.

We use as reference AirNet~\cite{Li_2022_CVPR}, IDR~\cite{zhang2023ingredient} and ADMS~\cite{park2023all}. We also consider the contemporary work PromptIR~\cite{potlapalli2023promptir}.  
The methods use different techniques to guide the blind model in the restoration process. For instance, an auxiliary model for degradation classification~\cite{Li_2022_CVPR, park2023all}, or multi-dimensional guidance vectors (also known as ``prompts")~\cite{potlapalli2023promptir, ma2023prores, liu2022ddrsr} that help the model to discriminate the different types of degradation in the image.

%Despite it is not the focus of this work, we acknowledge that \emph{real-world image super-resolution} is a related problem~\cite{liang2021swinir,zhang2021designing,luo2022learning,cornillere2019blind}, since the models aim to solve an inverse problem considering multiple degradations (blur, noise and downsampling).

\vspace{-3mm}
\paragraph{Text-guided Image Manipulation.}
In recent years, multiple methods have been proposed for text-to-image generation and text-based image editing works~\cite{brooks2023instructpix2pix, meng2021sdedit, ruiz2023dreambooth, kawar2023imagic, hertz2022prompt}. These models use text prompts to describe images or actions, and powerful diffusion-based models for generating the corresponding images. 
Our main reference is \instp~\cite{brooks2023instructpix2pix}, this method enables editing from {\em instructions} that tell the model what action to perform, as opposed to text labels, captions or descriptions of the input or output images. Therefore, the user can transmit what to do in natural written text, without requiring to provide further image descriptions or sample reference images. 

%%%%%%%%%%%%%%%%%%%%%%%%%%%%%%%%%%%%%%%%%%%%%%%%

\section{Image Restoration Following Instructions}
\label{sec:we}

We treat instruction-based image restoration as a supervised learning problem similar to previous works~\cite{brooks2023instructpix2pix}. First, we generate over 10000 prompts using GPT-4 based on our own sample instructions. We explain the creation of the prompt dataset in Sec.~\ref{sec:gen_prompt}. We then build a large paired training
dataset of prompts and degraded/clean images. Finally, we train our \we model, and we evaluate it on a wide variety of instructions including real human-written prompts. We explain our text encoder in Sec~\ref{sec:text_enc}, and our complete model in Sec.~\ref{sec:instructir}.

\subsection{Generating Prompts for Training}
\label{sec:gen_prompt}

\emph{\bf Why instructions?} Inspired by \instp~\cite{brooks2023instructpix2pix}, we adopt human written instructions as the mechanism of control for our model. There is no need for the user to provide additional information, such as example clean images, or descriptions of the visual content. Instructions offer a clear and expressive way to interact, enabling users to pinpoint the unpleasant effects (degradations) in the images. We also consider the language complexity, from ambiguous instructions (\eg ``fix my image'') to precise instructions (\eg ``remove the noise'').

\vspace{2mm}

\begin{wraptable}{r}{0.5\textwidth} % "l" for left, "5cm" for the width of the wraptable
\vspace{-9mm}
\caption{Examples of our curated GPT4-generated and real user prompts with varying language and domain expertise.}
\vspace{1mm}
  \label{tab:examples_prompt}
  \resizebox{\linewidth}{!}{%
  \begin{tabular}{ll}
    \toprule
   \bf{Degradation} & \bf{Prompts} \\
    \midrule
    \multirow{3}{*}{Denoising} & Can you clean the dots from my image? \\
    & Fix the grainy parts of this photo \\
    & Remove the noise from my picture \\
    \midrule
    \multirow{3}{*}{Deblurring} & Can you reduce the movement in the image? \\
    & My picture's not sharp, fix it \\
    & Deblur my picture, it's too fuzzy \\
    \midrule
    \multirow{3}{*}{Dehazing} & Can you make this picture clearer? \\
    & Help, my picture is all cloudy \\
    & Remove the fog from my photo \\
    \midrule
    \multirow{3}{*}{Deraining} & I want my photo to be clear, not rainy \\
    & Clear the rain from my picture \\
    & Remove the raindrops from my photo \\
    \midrule
    \multirow{3}{*}{Super-Res.} & Make my photo bigger and better \\
    & Add details to this image \\
    & Increase the resolution of this photo \\
    \midrule
    \multirow{3}{*}{Low-light} & The photo is too dark, improve exposure \\
    & Increase the illumination in this shot \\
    & My shot has very low dynamic range \\
    \midrule
    \multirow{3}{*}{Enhancement} & Make it pop! \\
    & Adjust the color balance for a natural look \\
    & Apply a cinematic color grade to the photo \\
    \midrule
    \multirow{2}{*}{General} & Fix my image please \\
    & make the image look better \\
    \bottomrule
  \end{tabular}
  }
\end{wraptable}

\begin{figure*}[!ht]
    \centering
    \includegraphics[trim={0.4cm 0 0 0},clip, width=\linewidth]{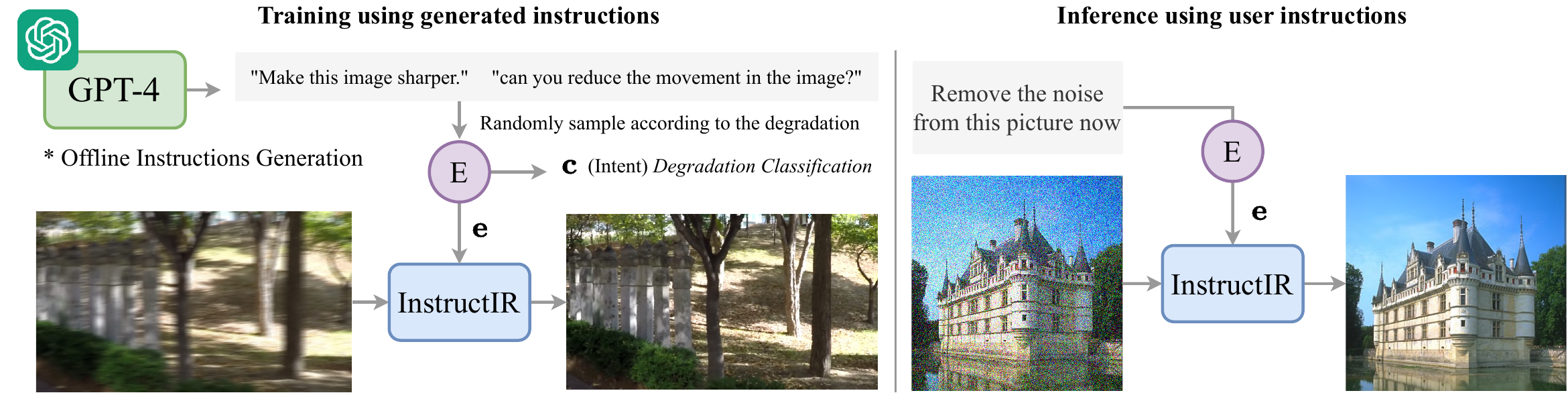}
    \caption{We train our blind image restoration models using common image datasets, and prompts generated using GPT-4, note that this is (self-)supervised learning. At inference time, our model generalizes to human-written instructions and restores (or enhances) the images.}
    \vspace{-1mm}
    \label{fig:diagram}
\end{figure*}

Handling free-form user prompts rather than fixed degradation-specific prompts increases the usability of our model for laypeople who lack domain expertise.
We thus want our model to be capable of understanding diverse prompts posed by users ``in-the-wild" \eg kids, adults, or photographers. 
To this end, we use a large language model (\ie, GPT-4) to create diverse requests that might be asked by users for the different degradations types.
We then filter those generated prompts to remove ambiguous or unclear prompts (\eg, \emph{``Make the image cleaner'', ``improve this image''}). 
Our final instructions set contains over 10000 different prompts in total, for 7 different tasks. We display some examples in Table~\ref{tab:examples_prompt}. As we show in Figure~\ref{fig:diagram} the prompts are sampled randomly depending on the input degradation.

\subsection{Text Encoder}
\label{sec:text_enc}

\paragraph{The Choice of the Text Encoder.}
A text encoder maps the user prompt to a fixed-size vector representation (a text embedding). The related methods for text-based image generation \cite{rombach2022stablediffusion} and manipulation \cite{brooks2023instructpix2pix,bai2023textir} often use the text encoder of a CLIP model \cite{radford2021clip} to encode user prompts as CLIP excels in visual prompts.
However, user prompts for degradation contain, in general, little to no visual content (\eg the use describes the degradation, not the image itself). %, therefore, the large CLIP encoders (with over 60 million parameters) are not suitable -- especially if we require efficiency. 
We opt, instead, to use a pure text-based sentence encoder \cite{reimers2019sbert}, that is, a smaller model trained to encode sentences in a semantically meaningful embedding space. Sentence encoders -- pre-trained with millions of examples -- are compact and fast in comparison to CLIP, while being able to encode the semantics of diverse user prompts. For instance, we use the \texttt{BGE-micro-v2} sentence transformer. We compare text encoders in the supplementary material.

\vspace{-4mm}
\paragraph{Fine-tuning the Text Encoder.}

We want to adapt the text encoder $\mathrm{E}$ for the restoration task to better encode the required information for the restoration model.
Training the full text encoder is likely to lead to overfitting on our small training set and lead to loss of generalization.
Instead, we freeze the text encoder and train a projection head on top:

\vspace{-2mm}
\begin{equation}
    \mathbf{e} = \mathrm{norm}(\mathbf{W} \cdot \mathrm{E}(t))
\end{equation}

where $t$ is the text, $\mathrm{E}(t)$ represents the raw text embedding, $\mathbf{W} \in \mathbb{R}^{d_t \times d_v}$ is a learned projection from the text dimension ($d_t$) to the input dimension for the restoration model ($d_v$), and $\mathrm{norm}$ is the l2-norm.
 
\begin{figure}[t]
    \centering
    \begin{subfigure}[b]{0.47\linewidth}
        \includegraphics[width=\linewidth]{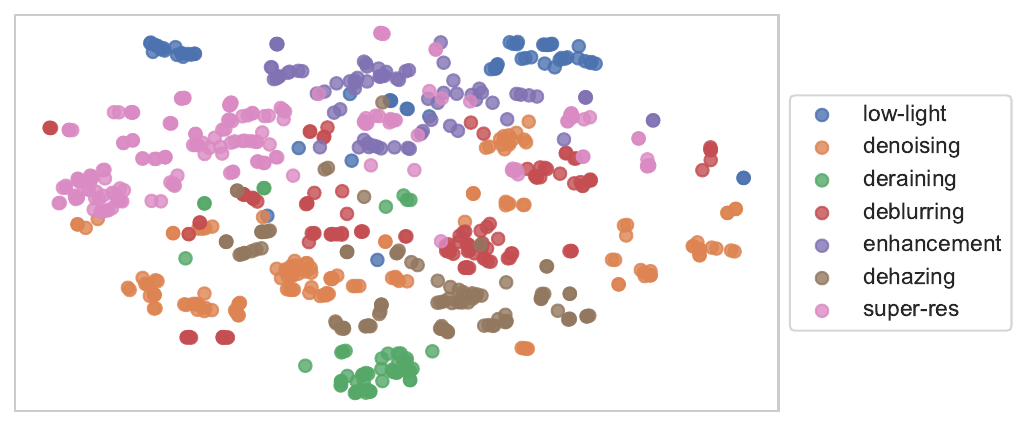}
        \caption{t-SNE of embeddings \emph{before} training \ie frozen text encoder}
    \label{fig:embeddings_tsne:encoder}
    \end{subfigure}
    \hspace{0.5cm} % Space between the images
    \begin{subfigure}[b]{0.47\linewidth}
        \includegraphics[width=\linewidth]{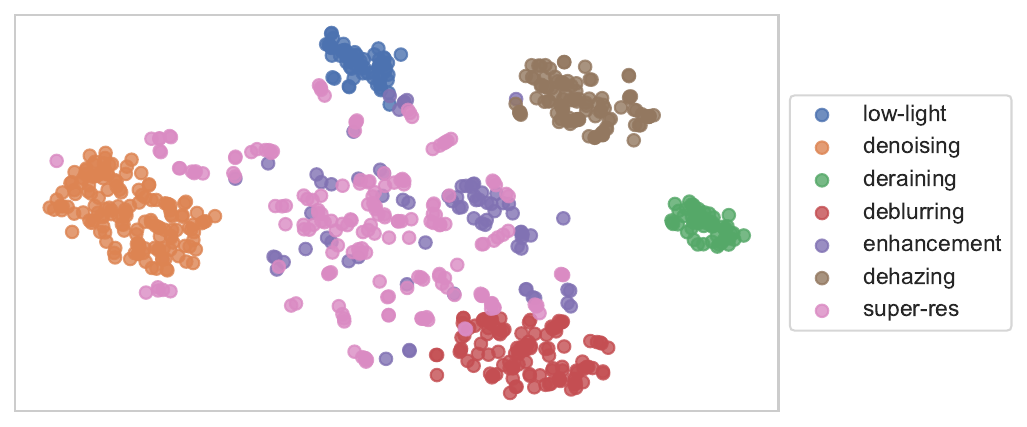}
        \caption{t-SNE of embeddings \emph{after} training our learned projection}
    \label{fig:embeddings_tsne:head}
    \end{subfigure}
    \caption{We show t-SNE plots of the text embeddings before/after training \we. Each dot represents a human instruction.
    }
    \vspace{-3mm}
    \label{fig:embeddings_tsne}
\end{figure}

Figure~\ref{fig:embeddings_tsne} shows that while the text encoder is capable out-of-the-box to cluster instructions to some extent (Figure~\ref{fig:embeddings_tsne:encoder}), our trained projection yields greatly improved clusters (Figure~\ref{fig:embeddings_tsne:head}). 
We distinguish clearly the clusters for deraining, denoising, dehazing, deblurring, and low-light image enhancement. The instructions for such tasks or degradations are very characteristic. Furthermore, we can appreciate that ``super-res" and ``enhancement" tasks are quite spread and between the previous ones, which matches the language logic. For instance \emph{``add details to this image"} could be used for enhancement, deblurring, or denoising.
In our experiments, $d_t\!=\!384$, $d_v\!=\!256$ and $\mathbf{W}$ is a linear layer. The representation $\mathbf{e}$ from the text encoder is shared across the blocks, and each block has a trainable projection $\mathbf{W}$.

\vspace{-4mm}
\paragraph{Intent Classification Loss.}
We propose a guidance loss on the text embedding $\mathbf{e}$ to improve training and interpretability. Using the degradation types as targets, we train a simple classification head $\mathcal{C}$ such that $\mathbf{c} = \mathcal{C} (\mathbf{e})$, where $\mathbf{c} \in \mathrm{R}^D$, being $D$ is the number of degradation classes. The classification head $\mathcal{C}$ is a simple two-layers MLP. Thus, we only need to train a projection layer $\mathbf{W}$ and a simple MLP to capture the natural language knowledge. This allows the text model to learn meaningful embeddings as we can appreciate in Figure~\ref{fig:embeddings_tsne}, not just guidance vectors for the main image processing model. 
We find that the model is able to classify accurately (\ie over 95\% accuracy) the underlying degradation in the user's prompt after a few epochs.

\subsection{InstructIR}
\label{sec:instructir}

Our method \we consists of an image model and a text encoder. We introduced our text encoder in Sec.~\ref{sec:text_enc}. We use NAFNet~\cite{chen2022simple} as the image model, an efficient image restoration model that follows a U-Net architecture~\cite{ronneberger2015unet}. To successfully learn multiple tasks using a single model, we use task routing techniques. Our framework for training and evaluation is illustrated in Figure~\ref{fig:diagram}.

\begin{wrapfigure}{r}{6cm}
    \vspace{-8mm}
    \centering
    \includegraphics[trim={1.2cm 0 0 0},clip,width=0.98\linewidth]{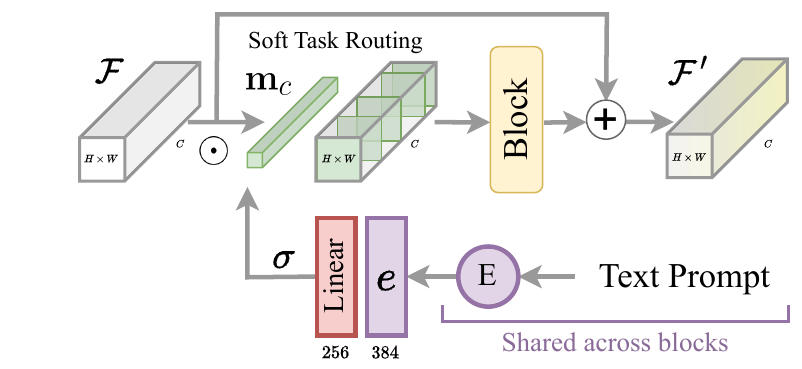}
    \vspace{-2mm}
    \caption{\emph{Instruction Condition Block (ICB)} using an approximation of task routing~\cite{strezoski2019manytask} for many-tasks learning (See Eq.~\ref{eq:block}). This mechanism allows the neural network to select and prioritize specific features depending on the instruction, similarly to a Mixture of Experts (MoE).
    }
    \label{fig:block}
    \vspace{-3mm}
\end{wrapfigure}

\paragraph{Text Guidance.}
The key aspect of \we is the integration of the encoded instruction as a mechanism of control for the image model. Inspired in \emph{task routing} for many-task learning~\cite{rosenbaum2017routing, strezoski2019manytask, ding2023task}, we propose an \emph{``Instruction Condition Block" (ICB)} to enable task-specific transformations within the model. 
Conventional task routing~\cite{strezoski2019manytask} applies task-specific binary masks to the channel features. Since our model does not know \emph{a-priori} the degradation, we cannot use this technique directly. 

Considering the image features $\mathcal{F}$, and the encoded instruction $\mathbf{e}$, we apply task routing as follows:

\vspace{-1mm}
\begin{equation}
    \mathcal{F'}_c = \mathrm{Block}(\mathcal{F}_c \odot \mathbf{m}_c ) + \mathcal{F}_c
\label{eq:block}
\end{equation}

where the mask $\mathbf{m}_c = \sigma(\mathbf{W_c} \cdot \mathbf{e})$ is produced using a linear layer $\mathbf{W_c}$ -- activated using the Sigmoid function -- to produce a set of weights depending on the text embedding $\mathbf{e}$. Thus, we obtain a $c$-dimensional per-channel (soft-)binary mask $\mathbf{m}_c$. As~\cite{strezoski2019manytask, he2019adafm}, task routing is applied as the channel-wise multiplication $\odot$ for masking features depending on the task. The conditioned features are further enhanced using a convolutional NAFBlock~\cite{chen2022simple} ($\mathrm{Block}$). We illustrate our task-routing ICB block in Figure~\ref{fig:block}. We use ``regular'' NAFBlocks~\cite{chen2022simple}, followed by ICBs to condition the features, at both encoder and decoder blocks. The formulation is $F^{l+1}\!=\!\mathrm{ICB}(\mathrm{Block}(F^{l}))$ where $l$ is the layer.
Although we do not condition explicitly the filters of the neural network, as in~\cite{strezoski2019manytask}, the mask allows the model to select the most relevant channels depending on the image information and the instruction. Note that this formulation enables differentiable feature masking, and certain interpretability \ie the features with high weights contribute the most to the restoration process. Indirectly, this also enforces to learn diverse filters and reduce sparsity~\cite{strezoski2019manytask, ding2023task}.

\vspace{-2mm}
\paragraph{Is \we a blind restoration model?}
The model does not use explicit information about the degradation in the image \eg noise profiles, blur kernels, or PSFs. Since our model infers the task (degradation) given the image and the instruction, we consider \we a \emph{blind} image restoration model. Similarly to previous works that use auxiliary image-based degradation classification~\cite{park2023all, Li_2022_CVPR}.

\section{Experimental Results}
\label{sec:results}

%We provide extensive qualitative results using benchmark images in Figures~\ref{fig:haze-quali},~\ref{fig:lol-comp},~\ref{fig:rain-comp}, and ~\ref{fig:blur-comp}. 
We evaluate our model on 9 well-known benchmarks for different image restoration tasks: image denoising, deblurring, deraining, dehazing, real low-light enhancement, and photo-realistic image enhancement. We present extensive quantitative results in Table~\ref{tab:results} and Table~\ref{tab:allinone}. We provide extensive comparisons with other all-in-one methods as well as task-specific methods. Our \emph{single} model successfully restores images considering different degradation types and levels. %\emph{We provide additional results and ablation studies in the supplementary material.}

\subsection{Implementation Details.}
Our \we model is end-to-end trainable. The image model does not require pre-training but we use a pre-trained sentence encoder as language model. 

\vspace{-2mm}
\paragraph{Text Encoder.} As we discussed in Sec.~\ref{sec:text_enc}, we only need to train the text embedding projection and classification head ($\approx\!100K$ parameters). We initialize the text encoder with \textsc{BGE-micro-v2}~\footnote{\url{https://huggingface.co/TaylorAI/bge-micro-v2}}, a distilled version of \textsc{BGE-small-en} \cite{xiao2023cpack}. The BGE encoders are BERT-like encoders \cite{devlin2019bert} pre-trained on large amounts of supervised and unsupervised data for general-purpose sentence encoding. The BGE-micro model is a 3-layer encoder with 17.3 million parameters, which we freeze during training. We also explore \textsc{all-MiniLM-L6-v2} and CLIP encoders, however, we concluded that small models prevent overfitting and provide the best performance while being fast. We provide the ablation study comparing the three text encoders in the supplementary material.

\vspace{-2mm}
\paragraph{Image Model.}
We use NAFNet~\cite{chen2022simple} as the image model backbone. The architecture consists of a 4-level encoder-decoder, with varying numbers of blocks at each level, specifically [2, 2, 4, 8] for the encoder, and [2, 2, 2, 2] for the decoder, from the level-1 to level-4 respectively. Between the encoder and decoder we use 4 middle blocks to enhance further the features. The decoder implements addition instead of concatenation for the skip connections. We use the \emph{Instruction Condition Block (ICB)} for task-routing~\cite{strezoski2019manytask} only in the encoder and decoder.

The model is optimized using the $\mathcal{L}_1$ loss between the ground-truth clean image and the restored one. Additionally, we use the cross-entropy loss $\mathcal{L}_{ce}$ for the intent classification head of the text encoder. We train using a batch size of 32 and AdamW~\cite{kingma2014adam} optimizer with learning rate $5e^{-4}$ for 500 epochs (approximately 1 day using a single NVIDIA A100). We also use cosine annealing learning rate decay. During training, we utilize cropped patches of size $256\times256$ as input, and we use random horizontal and vertical flips as augmentations. Since our model uses as input instruction-image pairs, given an image, and knowing its degradation, we randomly sample instructions from our prompt dataset ($>\!10$K samples). Our image model has only 16M parameters, and the learned text projection is just $100$k parameters (the language model is 17M parameters), thus, our model can be trained easily on standard GPUs, furthermore, the inference process also fits in low-computation budgets (\eg Google Colab T4 16Gb GPU).

\subsection{Datasets and Benchmarks} 

Following previous works~\cite{Li_2022_CVPR, zhang2023ingredient, potlapalli2023promptir}, we prepare the datasets for different restoration tasks, including real and synthetic datasets.

\vspace{-4mm}
\paragraph{Image denoising.} We use a combination of BSD400~\cite{amfm_pami2011_bsd500} and WED~\cite{ma2016waterloo_wed} datasets for training. This combined training set contains $\approx\!5000$ images. Using as reference the clean images in the dataset, we generate the noisy images by adding Gaussian noise with different noise levels $\sigma \in \{15,25,50\}$.  We test the models on the well-known BSD68~\cite{martin2001database_bsd} and Urban100~\cite{huang2015single} datasets.

\vspace{-4mm}
\paragraph{Image deraining.} We use the Rain100L~\cite{yang2020learning} dataset, which consists of 200 clean-rainy image pairs for training, and 100 pairs for testing. 

\vspace{-4mm}
\paragraph{Image dehazing.} We utilize the Reside (outdoor) SOTS~\cite{li2018benchmarking} dataset, which contains $\approx\!72$K training images. However, many images are low-quality and unrealistic, thus, we filtered the dataset and selected a random set of 2000 images -- also to avoid imbalance \emph{w.r.t} the other tasks. We use the standard \emph{outdoor} test set of 500 images.

\vspace{-4mm}
\paragraph{Image deblurring.} We use the GoPro dataset for motion deblurring~\cite{gopro2017} which consists of 2103 images for training, and 1111 for testing.

\vspace{-4mm}
\paragraph{Real-world Low-light Image Enhancement.} We use the LOL~\cite{Chen2018Retinex} dataset~(v1), which contains real-case low/normal-light image pairs. We adopt its official split of 485 training images and 15 testing images.

\vspace{-4mm}
\paragraph{Real-world Image Enhancement.} Extending previous works, we also study photo-realistic image enhancement using the MIT5K DSLR dataset~\cite{fivek}. We use 1000 images for training, and the standard split of 500 images for testing (as in~\cite{tu2022maxim}).

\vspace{2mm}
\noindent Finally, as previous works~\cite{Li_2022_CVPR, zhang2023ingredient, potlapalli2023promptir}, we combine all the aforementioned training datasets, and we train our unified model for all-in-one restoration. Note that we do not include more \emph{real-world datasets} because previous works do not provide results (or models) for those. Moreover, previous works were limited to synthetic data, in contrast, \we also tackles real-world image enhancement.

%%%%%%%%%%%%%%%%%%%%%%%%%%%%%%%%%%%%%%%%%%%%%%%%%%%%%%%%%%%%%%%%%%
%%% 5D TASKS TABLE
\begin{table*}[t]
\centering
\caption{Quantitative results on \emph{\bf five restoration tasks (\textcolor{purple}{\bf 5D})} with \emph{state-of-the-art} general image restoration and
all-in-one methods. We highlight the \colorbox{LightCyan}{reference model \emph{without} text (image only)}, the \colorbox{lyellow}{best overall results}, and the \colorbox{lgreen}{second best results}. We also present the ablation study of our \emph{multi-task variants} (from 5 to 7 tasks --- 5D, 6D, 7D). This table is based on Zhang \emph{et al.} IDR~\cite{zhang2023ingredient}.}
\vspace{-1mm}
\label{tab:results}
%\begin{tabular}{@{}lS[table-format=2.2]*{12}{S[table-format=2.2,table-space-text-post=***]}l@{}}
% \footnotesize
\resizebox{\linewidth}{!}{%
\begin{tabularx}{1.45\linewidth}{r c*{12}{c}}
\toprule
& \multicolumn{2}{c}{\bf Deraining} &  \multicolumn{2}{c}{\bf Dehazing} & \multicolumn{2}{c}{\bf Denoising} & \multicolumn{2}{c}{\bf Deblurring} & \multicolumn{2}{c}{\bf Low-light Enh.} & & & \\
\textbf{Methods} & \multicolumn{2}{c}{Rain100L~\cite{yang2020learning}} & \multicolumn{2}{c}{SOTS~\cite{li2018benchmarking}} & \multicolumn{2}{c}{BSD68~\cite{martin2001database_bsd}} & \multicolumn{2}{c}{GoPro~\cite{gopro2017}} & \multicolumn{2}{c}{LOL~\cite{Chen2018Retinex}} & \multicolumn{2}{c}{\bf Average} & \textbf{Params} \\
\cmidrule(lr){2-3} \cmidrule(lr){4-5} \cmidrule(lr){6-7} \cmidrule(lr){8-9} \cmidrule(lr){10-11} \cmidrule(lr){12-13}
 & {PSNR↑} & {SSIM↑} & {PSNR↑} & {SSIM↑} & {PSNR↑} & {SSIM↑} & {PSNR↑} & {SSIM↑} & {PSNR↑} & {SSIM↑} & {PSNR↑} & {SSIM↑} & (M) \\
\midrule
HINet~\cite{chen2021hinet}       & 35.67 & 0.969 & 24.74 & 0.937 & 31.00 & 0.881 & 26.12 & 0.788 & 19.47 & 0.800 & 27.40 & 0.875 & 88.67 \\
DGUNet~\cite{mou2022deep}     & 36.62 & 0.971 & 24.78 & 0.940 & 31.10 & 0.883 & 27.25 & 0.837 & 21.87 & 0.823 & 28.32 & 0.891 & 17.33 \\
MIRNetV2~\cite{zamir2020mirnet}   & 33.89 & 0.954 & 24.03 & 0.927 & 30.97 & 0.881 & 26.30 & 0.799 & 21.52 & 0.815 & 27.34 & 0.875 & 5.86  \\
SwinIR~\cite{liang2021swinir}     & 30.78 & 0.923 & 21.50 & 0.891 & 30.59 & 0.868 & 24.52 & 0.773 & 17.81 & 0.723 & 25.04 & 0.835 & 0.91  \\
Restormer~\cite{restormer}  & 34.81 & 0.962 & 24.09 & 0.927 & 31.49 & 0.884 & 27.22 & 0.829 & 20.41 & 0.806 & 27.60 & 0.881 & 26.13 \\
NAFNet~\cite{chen2022simple}      & 35.56 & 0.967 & 25.23 & 0.939 & 31.02 & 0.883 & 26.53 & 0.808 & 20.49 & 0.809 & 27.76 & 0.881 & 17.11 \\
\midrule
DL~\cite{fan2019general}        & 21.96 & 0.762 & 20.54 & 0.826 & 23.09 & 0.745 & 19.86 & 0.672 & 19.83 & 0.712 & 21.05 & 0.743 & 2.09  \\
Transweather~\cite{valanarasu2022transweather} & 29.43 & 0.905 & 21.32 & 0.885 & 29.00 & 0.841 & 25.12 & 0.757 & 21.21 & 0.792 & 25.22 & 0.836 & 37.93 \\
TAPE~\cite{liu2022tape}       & 29.67 & 0.904 & 22.16 & 0.861 & 30.18 & 0.855 & 24.47 & 0.763 & 18.97 & 0.621 & 25.09 & 0.801 & 1.07  \\
AirNet~\cite{Li_2022_CVPR}     & 32.98 & 0.951 & 21.04 & 0.884 & 30.91 & 0.882 & 24.35 & 0.781 & 18.18 & 0.735 & 25.49 & 0.846 & 8.93  \\
\rowcolor{LightCyan} \we~w/o text & 35.58 & 0.967 & 25.20 & 0.938 & 31.09 & 0.883 & 26.65 & 0.810 & 20.70 & 0.820 & 27.84 & 0.884 & 17.11 \\
\rowcolor{lgreen}IDR~\cite{zhang2023ingredient} 
& 35.63 & 0.965 & 25.24 & 0.943 & 31.60 & 0.887 & 27.87 & 0.846 & 21.34 & 0.826 & \underline{28.34} & \underline{0.893} & 15.34 \\
\midrule
\rowcolor{lyellow} \we-\textcolor{purple}{\bf 5D}
& 36.84 & 0.973 & 27.10 & 0.956 & 31.40 & 0.887 & 29.40 & 0.886 & 23.00 & 0.836 & \textbf{29.55} & \textbf{0.907} & 15.8 \\
\we-{\textbf{\textcolor{sblue}{6D}}}            
& 36.80 & 0.973 & 27.00 & 0.951 & 31.39 & 0.888 & 29.73 & 0.892 & 22.83 & 0.836 & \textbf{29.55} & \textbf{0.908} & 15.8 \\
\we-{\bf 7D}            
& 36.75 & 0.972 & 26.90 & 0.952 & 31.37 & 0.887 & 29.70 & 0.892 & 22.81 & 0.836 & 29.50 & 0.907 & 15.8 \\
\bottomrule
\end{tabularx}
}
\end{table*}

%%%%%%%%%%%%%%%%%%%%%%%%%%%%%%%%%%%%%%%%%%%%%%%%%%%%%%%%%%
%%% 3D TASKS TABLE
\begin{table*}[!ht]
  \centering
  % \footnotesize
  \caption{Comparisons of all-in-one restoration models for \emph{\bf 3 restoration tasks (\textcolor{lpurple}{3D})}. We also show an ablation study for image denoising -the fundamental inverse problem- considering different noise levels. We report PSNR/SSIM metrics. Table based on~\cite{potlapalli2023promptir}.}
  \label{tab:allinone}
  \vspace{-1mm}
  \resizebox{0.9\textwidth}{!}{
  \begin{tabular}{rcccccc}
    \toprule
    \textbf{Methods}  & \textbf{Dehazing} & \textbf{Deraining} &  \multicolumn{3}{c}{\textbf{Denoising ablation study (BSD68~\cite{martin2001database_bsd})}} &
    \textbf{Average} \\
    %\vspace{0.5pt}
    & SOTS~\cite{li2018benchmarking}& Rain100L~\cite{fan2019general}& $\sigma = 15$ & $\sigma = 25$ & $\sigma = 50$ & \\
    \midrule
    BRDNet~\cite{tian2020BRDnet} & 23.23/0.895 & 27.42/0.895 & 32.26/0.898 & 29.76/0.836 &  26.34/0.836 & 27.80/0.843 \\
    LPNet~\cite{gao2019dynamic} & 20.84/0.828 & 24.88/0.784 &  26.47/0.778 & 24.77/0.748 & 21.26/0.552 & 23.64/0.738  \\
    FDGAN~\cite{dong2020fd} & 24.71/0.924 & 29.89/0.933 & 30.25/0.910 & 28.81/0.868 & 26.43/0.776 & 28.02/0.883 \\
    MPRNet~\cite{Zamir_2021_CVPR_mprnet} & 25.28/0.954 & 33.57/0.954 & 33.54/0.927 & 30.89/0.880 & 27.56/0.779 & 30.17/0.899 \\
    DL\cite{fan2019general} & 26.92/0.931 & 32.62/0.931 & 33.05/0.914 & 30.41/0.861 & 26.90/0.740 & 29.98/0.875 \\
    AirNet~\cite{Li_2022_CVPR} & {27.94}/{0.962} &{34.90}/{0.967} & {33.92}/{0.933} & {31.26}/{0.888} & {28.00}/{0.797} & {31.20}/{0.910} \\
    \rowcolor{lgreen} PromptIR~\cite{potlapalli2023promptir} 
    & \textbf{30.58}/\textbf{0.974} & \underline{36.37}/\underline{0.972} & \underline{33.98}/\underline{0.933} & \underline{31.31}/\underline{0.888} & \underline{28.06}/\underline{0.799} & \underline{32.06}/\underline{0.913} \\
    \rowcolor{lyellow} \we-\textcolor{lpurple}{\bf 3D} 
    & \underline{30.22}/\underline{0.959} & \textbf{37.98}/\textbf{0.978} & \textbf{34.15}/\textbf{0.933} & \textbf{31.52}/\textbf{0.890} & \textbf{28.30}/\textbf{0.804} & \textbf{32.43}/\textbf{0.913} \\
    \we-\textcolor{purple}{\bf 5D} 
    & 27.10/0.956 & 36.84/0.973 & 34.00/0.931 & 31.40/0.887   & 28.15/0.798 & 31.50/0.909 \\
    \rowcolor{LightCyan} \we~w/o text & {26.84}/{0.948} &{34.02}/{0.960} & {33.70}/{0.929} & {30.94}/{0.882} & {27.78}/{0.780} & {30.65}/{0.900} \\
    \bottomrule
  \end{tabular}}
\end{table*}
%%%%%%%%%%%%%%%%%%%%%%%%%%%%%%%%%%%%%%%%%%%%%%%%%%%%%%%%%%

\subsection{Multiple Degradation Results}

We define two initial setups for multi-task restoration:
\begin{itemize}
    \itemsep0em 
    \item \textcolor{lpurple}{\bf 3D} for \emph{three-degradation} models such as AirNet~\cite{Li_2022_CVPR}, these tackle image denoising, dehazing and deraining.
    
    \item \textcolor{purple}{\bf 5D} for \emph{five-degradation} models, considering image denoising, deblurring, dehazing, deraining and low-light image enhancement as in~\cite{zhang2023ingredient}. 
\end{itemize}

%\vspace{2mm}

In Table~\ref{tab:results}, we show the performance of \textcolor{purple}{\bf 5D} models. Following Zhang \emph{et al.}~\cite{zhang2023ingredient}, we compare \we with several \sota methods for general image restoration~\cite{restormer, chen2022simple, chen2021hinet, liang2021swinir, zamir2020mirnet}, and all-in-one image restoration methods~\cite{zhang2023ingredient, Li_2022_CVPR, valanarasu2022transweather, fan2019general, liu2022tape}. We can observe that our simple image model (just 16M parameters) can tackle successfully at least five different tasks thanks to the instruction-based guidance and achieves the most competitive results. 
In Table~\ref{tab:allinone} we can appreciate a similar behavior, when the number of tasks is just three (\threed), our model improves further in terms of reconstruction performance. 

Based on these results, we pose the following question: \emph{How many tasks can we tackle using a single model without losing too much performance?} To answer this, we propose the \textbf{\textcolor{sblue}{6D}} and \textbf{7D} variants. For the \textbf{\textcolor{sblue}{6D}} variant, we fine-tune the original \textcolor{purple}{\bf 5D} to consider also super-resolution as sixth task. Finally, our \textbf{7D} model includes all previous tasks, and additionally image enhancement (MIT5K photo retouching).
We show the performance of these two variants in Table~\ref{tab:results}. %We extend this study in the supplementary material.

%%%%%%%%%%%%%%%%%%%%%%%%%%%%%%%%%%%%%%%%%%%%%%%%%%%%%%%%%%
%% LANGUAGE ABLATION
\begin{wraptable}{r}{0.5\textwidth}
    \vspace{-8mm}
    \caption{Ablation study on the \emph{sensitivity of instructions}. We report PSNR/SSIM metrics for each task using our \textcolor{purple}{\bf 5D} base model. We repeat the evaluation on each test set 10 times, each time we sample different prompts for each image, and we report the average results. The ``Real Users~$\dagger$" in this study are amateur photographers, thus, the instructions were very precise.}
    \label{tab:a-prompt}
    \resizebox{\linewidth}{!}{
    \begin{tabular}{l c c c c}
    \toprule
    \textbf{Language Level}     & \textbf{Deraining} & \textbf{Denoising} & \textbf{Deblurring} & \textbf{LOL}  \\
    \midrule
    \rowcolor{lyellow}Basic \& Precise                            & 36.84/0.973 & 31.40/0.887 & 29.47/0.887  & 23.00/0.836 \\
    Basic \& Ambiguous    & 36.24/0.970 & 31.35/0.887 & 29.21/0.885  & 21.85/0.827 \\
    Real Users~$\dagger$            & 36.84/0.973 & 31.40/0.887 & 29.47/0.887  & 23.00/0.836 \\
    \bottomrule
    \end{tabular}
    }
    %\vspace{-3mm}
\end{wraptable}%
%%%%%%%%%%%%%%%%%%%%%%%%%%%%%%%%%%%%%%%%%%%%%%%%%%%%%%%%%%

\begin{figure*}[t]
    \centering
    \setlength{\tabcolsep}{1pt}
    \begin{tabular}{c c c c}
         %\midrule
         %\rowcolor{lyellow} \multicolumn{4}{c}{Instruction: \emph{``Reduce the noise in this photo"} -- Basic \& Precise}\\
         %\midrule
         \includegraphics[trim={0 0 0 0},clip, width=0.24\linewidth]{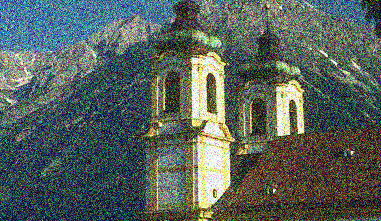} & 
         \includegraphics[trim={0 0 0 0},clip, width=0.24\linewidth]{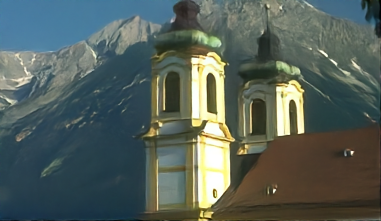} & 
         \includegraphics[trim={0 0 0 0},clip, width=0.24\linewidth]{figs/inst_var/ns_ours_1.png} & 
         \includegraphics[trim={0 0 0 0},clip, width=0.24\linewidth]{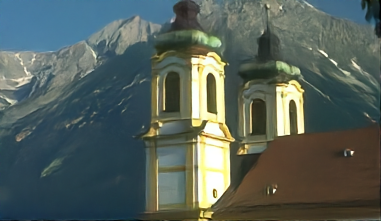} \\
         Input & \emph{``Clean up my image,}  &  \emph{``Get rid of the grain}  & \emph{``Remove the strange} \\
         & \emph{it's too fuzzy."} & \emph{in my photo"} & \emph{spots on my photo"} \\
         \includegraphics[trim={0 0 0 0},clip, width=0.24\linewidth]{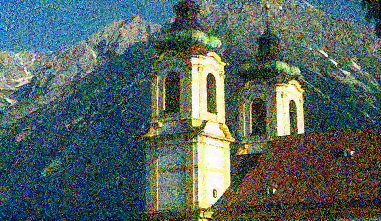} & 
         \includegraphics[trim={0 0 0 0},clip, width=0.24\linewidth]{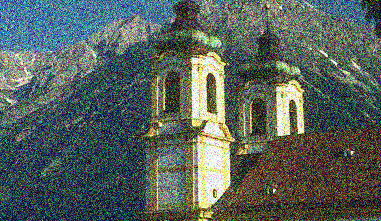} & 
         \includegraphics[trim={0 0 0 0},clip, width=0.24\linewidth]{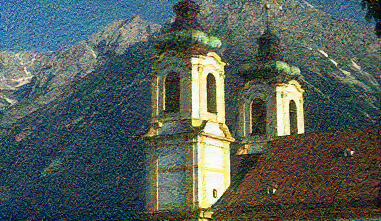} & 
         \includegraphics[trim={0 0 0 0},clip, width=0.24\linewidth]{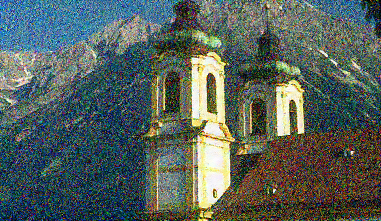} \\
         \emph{``Retouch this image}  & \emph{``Reduce the motion}  &  \emph{``Please get rid of}  & \emph{``Reduce the fog in}  \\
         \emph{and improve colors"} & \emph{in this shot"} & \emph{the raindrops"} & \emph{this landmark"} \\
         
    \end{tabular}
    \caption{\textbf{Adversarial and OOD samples for Instruction-based Restoration}. \we understands a wide range of instructions for a given task (first row). Given an \emph{adversarial or out-of-distribution instruction} (second row), the model does not modify the image notably (\ie performs the identity) --we did not enforce this during training--.}
    \label{fig:comp-instr}
\end{figure*}

\begin{figure*}[!ht]
    \centering
    \setlength{\tabcolsep}{1pt}
    \resizebox{\linewidth}{!}{
    \begin{tabular}{c c c}
         %\midrule
         %\rowcolor{lyellow} \multicolumn{4}{c}{Instruction: \emph{``Reduce the noise in this photo"} -- Basic \& Precise}\\
         %\midrule
         \fbox{\includegraphics[trim={0 1cm 0 0},clip, width=0.33\linewidth]{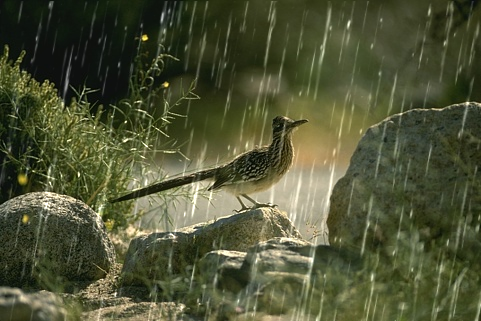}} & 
         \includegraphics[trim={0 1cm 0 0},clip, width=0.33\linewidth]{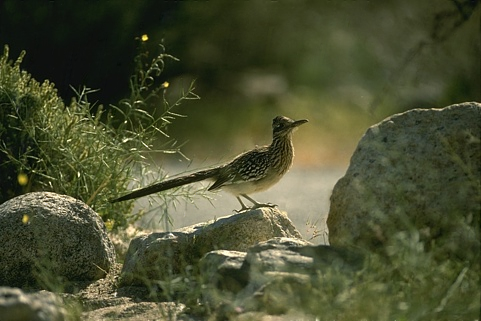} & 
         \includegraphics[trim={0 1cm 0 0},clip, width=0.33\linewidth]{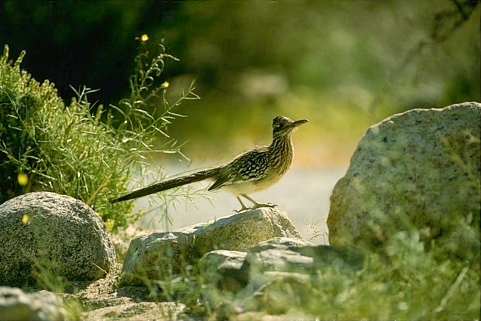} \\
         Input & \multicolumn{2}{c}{\scriptsize{\emph{(1)~``Clear the rain from my picture" $\longrightarrow$} \emph{(2)~``Make it pop"}}
         } \\
         \includegraphics[trim={0 1cm 0 0},clip, width=0.33\linewidth]{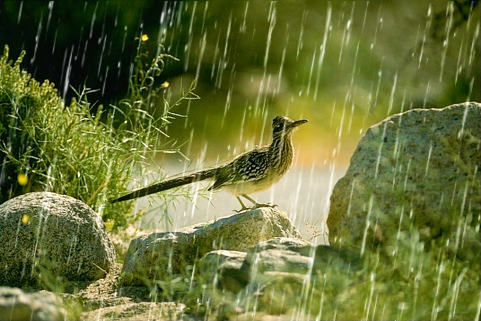} & 
         \includegraphics[trim={0 1cm 0 0},clip, width=0.33\linewidth]{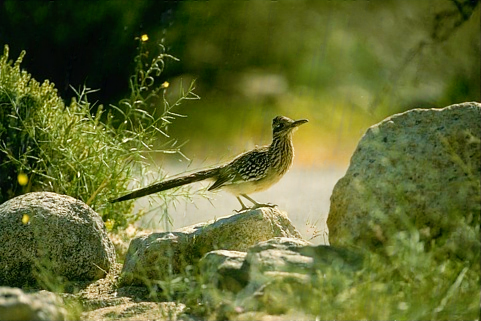} &
         \includegraphics[trim={0 1cm 0 0},clip, width=0.33\linewidth]{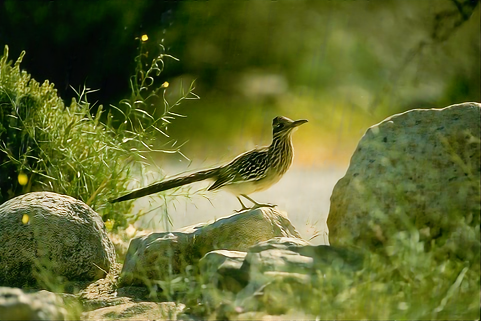} \\
         \multicolumn{3}{c}{\scriptsize{
         \emph{(1)~``Retouch it as a photographer"} $\longrightarrow$ ``Can you remove the raindrops?" $\longrightarrow$ ``Increase the resolution"
         }}  \\
         %
         %\includegraphics[trim={0 0 0 1cm},clip, width=0.33\linewidth]{figs/inst_var/gopro-min.png} & 
         %\includegraphics[trim={0 0 0 1cm},clip, width=0.33\linewidth]{figs/inst_var/gopro_1-min.png} & 
         %\includegraphics[trim={0 0 0 1cm},clip, width=0.33\linewidth]{figs/inst_var/gopro_2-min.png} \\
         %Input & \emph{``Reduce the movement in this pic"}  &  \emph{``Enhance the quality of my photo."}  \\
         %
         %\\
         \fbox{\includegraphics[trim={0 0 0 1cm},clip, width=0.33\linewidth]{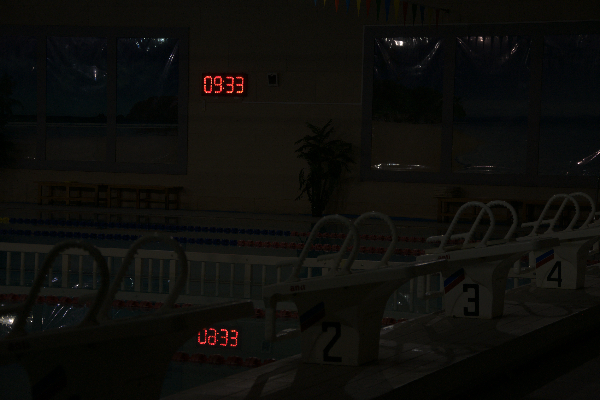}} & 
         \includegraphics[trim={0 0 0 1cm},clip, width=0.33\linewidth]{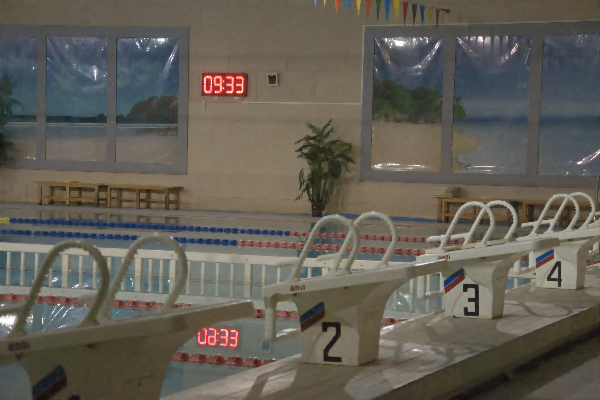} & 
         \includegraphics[trim={0 0 0 1cm},clip, width=0.33\linewidth]{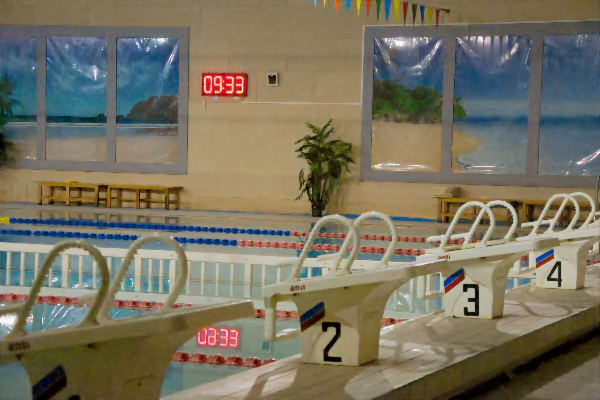} \\
         Input & \multicolumn{2}{c}{\scriptsize{\emph{(1)~``My image is too dark, fix it" $\longrightarrow$} \emph{(2)~``Apply a tonemap"}}
         } \\
         %Input & \emph{(1)``My image is too dark, can you fix it?"}  &  \emph{(2)``Apply tone-mapping to the photo"}  \\
    \end{tabular}
    }
    \caption{\textbf{Control via instructions}. We can prompt multiple instructions (in sequence) to restore and enhance the images. This provides additional \emph{control}. We show two examples of multiple instructions applied to the ``Input" image -from left to right-.}
    \label{fig:multi-instr}
\end{figure*}

\paragraph{Test Instructions.} \we requires as input the degraded image \underline{and} the human-written instruction. Therefore, we also prepare a test set of prompts \ie instruction-image test pairs. The performance of \we depends on the ambiguity and precision of the instruction. We provide the ablation study in Table~\ref{tab:a-prompt}. \we is quite robust to more/less detailed instructions. However, it is still limited with ambiguous instructions such as \emph{``enhance this image"}. We show diverse instructions in Figures~\ref{fig:comp-instr} and~\ref{fig:multi-instr}.

%%%%%%%%%%%%%%%%%%%%%%%%%%%%%%%%%%%%%%%%%%%%%%%%%%%%%%%%%%
\section{Multi-Task Discussion and Study}

%Previous works tackle 3 or 5 tasks (degradations). Considering the benefits of instruction-based restoration, we study the \textbf{\textcolor{sblue}{6D}} and \textbf{7D} variants described in the paper \emph{Sec. 4.3}.
%For the \textbf{\textcolor{sblue}{6D}} variant, we fine-tune the original \textcolor{purple}{\bf 5D} to add also super-resolution as sixth task. Moreover, our \textbf{7D} model includes all previous tasks, and additionally photo-realistic image enhancement (MIT5K photo retouching~\cite{fivek}).

\paragraph{How does 6D work?} Besides the 5 basic tasks -as previous works-, we include single image super-resolution (SISR). For this, we include as training data the DIV2K~\cite{Agustsson_2017_CVPR_Workshops_div2k}. Since our model does not perform upsampling, we use the Bicubic degradation model~\cite{Agustsson_2017_CVPR_Workshops_div2k, dong2015image} for generating the low-resolution images (LR), and the upsampled versions (HR) that are fed into our model to enhance them. Adding this extra task increases the performance on deblurring --a related degradation--, without harming notably the performance on the other tasks. %However, the performance on SR benchmarks is far from classical super-resolution methods~\cite{Agustsson_2017_CVPR_Workshops_div2k, liang2021swinir}.

%%%%%%%%%%%%%%%%%%%%%%%%%%%%%%%%%%%%%%%%%%%%%%%%%%%%%%%%
%% IMAGE ENHANCEMENT
\begin{wraptable}{r}{0.4\textwidth}
    \centering
    \vspace{-12mm}
    \caption{\textbf{Real Image Enhancement} results on MIT5K~\cite{fivek}.}
    \label{tab:fivek}
    \resizebox{\linewidth}{!}{
    \begin{tabular}{r c c c}
         \toprule
         \textbf{Method} & \textbf{PSNR}~$\uparrow$ & \textbf{SSIM}~$\uparrow$ & \textbf{$\Delta\!E_{ab}~\downarrow$} \\
         \midrule
         UPE~\cite{wang2019underexposed5k}  & 21.88 & 0.853 & 10.80 \\
         DPE~\cite{gharbi2017deep5k}        & 23.75 & 0.908 & 9.34 \\
         HDRNet~\cite{chen2018deep5k}       & 24.32 & \underline{0.912} & 8.49 \\
         3DLUT~\cite{zeng2020learning-3dlut} & \textbf{25.21} & \textbf{0.922} & \textbf{7.61} \\
         \rowcolor{lyellow} \we-{\bf 7D}    & \underline{24.65} & 0.900 & \underline{8.20} \\
         \bottomrule
    \end{tabular}
    }
\end{wraptable}
%%%%%%%%%%%%%%%%%%%%%%%%%%%%%%%%%%%%%%%%%%%%%%%%%%%%%%%%

\paragraph{How does 7D work?} Finally, if we add real image enhancement --a task not related to the previous ones \ie inverse problems-- the performance on the restoration tasks decays slightly. However, the model still achieves \sota results. Moreover, as we show in Table~\ref{tab:fivek}, the performance on this task using the MIT5K~\cite{fivek} dataset is notable, while keeping the performance on the other tasks. %We achieve similar performance to classical task-specific methods.

%%%%%%%%%%%%%%%%%%%%%%%%%%%%%%%%%%%%%%%%%%%%%%%%%%%%%%%%
%% SUMMARY TASKS ABLATION
\begin{wraptable}{r}{0.5\textwidth}
    \vspace{-5mm}
    \caption{\textbf{Summary ablation study} on the multi-task variants of \we that tackle from 3 to 7 tasks.}
    \label{tab:summary}
    \resizebox{\linewidth}{!}{
    \begin{tabular}{c c c c c}
    \toprule
    \textbf{Tasks}    & \textbf{Rain} & \textbf{Noise ($\sigma15$)} & \textbf{Blur} & \textbf{LOL}  \\
    \midrule
    \threed   & 37.98/0.978 & 31.52/0.890 & -  &  - \\
    \fived    & 36.84/0.973 & 31.40/0.887 & 29.40/0.886  & 23.00/0.836 \\
    \sixd     & 36.80 0.973 & 31.39 0.888 & 29.73/0.892  & 22.83 0.836 \\
    \sevend   & 36.75 0.972 & 31.37 0.887 & 29.70/0.892  & 22.81 0.836 \\
    \bottomrule
    \end{tabular}
    }
    \vspace{-17mm}
\end{wraptable}%
%%%%%%%%%%%%%%%%%%%%%%%%%%%%%%%%%%%%%%%%%%%%%%%%%%%%%%%%

\noindent We \textbf{summarize} the multi-task ablation study in Table~\ref{tab:summary}. Our model can tackle multiple tasks without losing performance notably thanks to the instruction-based task routing.

%\vspace{10mm}
\paragraph{Comparison with Task-specific Methods}

Our main goal is to design a powerful all-in-one model, thus, \we was not designed to be trained for a particular degradation. Nevertheless, we also compare \we with task-specific methods \ie models tailored and trained for specific tasks.

We compare with task-specific methods for real-world photography enhancement in Table~\ref{tab:fivek}, and for real-world low-light image enhancement in Table~\ref{tab:lol}. 

%We provide extensive comparisons for image denoising in Table~\ref{tab:noise}. Also, in Table~\ref{tab:gopro-sots} we show comparisons with classical methods for deblurring and dehazing. %Our multi-task method is better than most task-specific methods, yet it is still not better than SOTA. 

%%%%%%%%%%%%%%%%%%%%%%%%%%%%%%%%%%%%%%%%%%%%%%%%%%%
%%% NOISE INSTRUCT
%%%%%%%%%%%%%%%%%%%%%%%%%%%%%%%%%%%%%%%%%%%%%%%%%%%
%%% LOL BENCHMARK
\begin{table*}[t]
\centering
\caption{Quantitative comparisons with \sota methods on the \textbf{LOL~\cite{Chen2018Retinex}} dataset for Real-World Low-light Enhancement. Note that \we-{\textbf{{7D}}} is a multi-task method, while the other methods are task-specific. Table based on~\cite{wang2023low}.}
\label{tab:lol}
\vspace{-2mm}
\resizebox{\linewidth}{!}{
\begin{tabular}{cccccccccccc}
    \toprule
    \textbf{Method} & 
    \makecell[c]{LPNet \\\cite{li-tmm20-luminance}} &
    \makecell[c]{URetinex \\ -Net\cite{wu-cvpr22-uretinex}} & \makecell[c]{DeepLPF \\ \cite{moran-cvpr20-deeplpf}} & \makecell[c]{SCI \\ \cite{ma-cvpr22-toward}} & \makecell[c]{LIME \\ \cite{guo-tip16-lime}} & \makecell[c]{MF \\ \cite{fu-sp16-a}} & \makecell[c]{NPE \\ \cite{wang-tip13-naturalness}} & \makecell[c]{SRIE \\ \cite{fu-cvpr16-a}} & \makecell[c]{SDD \\ \cite{hao-tmm20-low}} & \makecell[c]{CDEF \\\cite{lei-tmm22-low}} & \makecell[c]{\cellcolor{lyellow} \textbf{\we} \\ \cellcolor{lyellow} \emph{Ours}} \\
    \midrule
    \textbf{PSNR}~$\uparrow$ & 21.46 & 21.32 & 15.28 & 15.80 & 16.76 & 16.96 & 16.96 & 11.86 & 13.34 & 16.33 & \cellcolor{lyellow} \underline{22.81} \\
    \textbf{SSIM}~$\uparrow$ & 0.802 & 0.835 & 0.473 & 0.527 & 0.444 & 0.505 & 0.481 & 0.493 & 0.635 & 0.583 & \cellcolor{lyellow}\underline{0.836} \\
    \midrule
    \midrule
    \textbf{Method} & \makecell[c]{DRBN \\ \cite{yang-tip21-band}} & \makecell[c]{KinD \\ \cite{zhang-acmmm19-kindling}} & \makecell[c]{RUAS \\ \cite{liu-cvpr21-retinex}} & \makecell[c]{FIDE \\ \cite{xu-cvpr20-learning}} & \makecell[c]{EG \\ \cite{jiang-tip21-enlightengan}} & \makecell[c]{MS-RDN \\ \cite{yang-tip21-sparse}} & \makecell[c]{Retinex \\ -Net\cite{Chen2018Retinex}} & \makecell[c]{MIRNet \\ \cite{zamir2020mirnet}} & \makecell[c]{IPT \\\cite{chen2021IPT}} & \makecell[c]{Uformer \\\cite{wang2021uformer}} & \makecell[c]{\textbf{IAGC} \\\cite{wang2023low}}  \\
    \midrule
    \textbf{PSNR}~$\uparrow$ & 20.13 & 20.87 & 18.23 & 18.27 & 17.48 & 17.20 & 16.77 & 24.14 & 16.27 & 16.36 & \textbf{24.53}\\
    \textbf{SSIM}~$\uparrow$ & 0.830 & 0.800 & 0.720 & 0.665 & 0.650 & 0.640 & 0.560 & 0.830 & 0.504 & 0.507 & \textbf{0.842}\\
    \bottomrule
\end{tabular}
}
\vspace{-5mm}
\end{table*}
%%%%%%%%%%%%%%%%%%%%%%%%%%%%%%%%%%%%%%%%%%%%%%%%%%%%%%%%
%%% MULTI-INSTRUCT
%%%%%%%%%%%%%%%%%%%%%%%%%%%%%%%%%%%%%%%%%%%%%%%%%%%%%%%%
%%% COMPARISON INSTRUCTPIX2PIX
\begin{figure*}[!ht]
    \centering
    \setlength{\tabcolsep}{1pt}
    \begin{tabular}{c c c c }
         \includegraphics[trim={0 2cm 0 0},clip, width=0.22\linewidth]{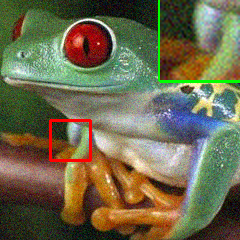} & 
         \includegraphics[trim={0 2cm 0 0},clip, width=0.22\linewidth]{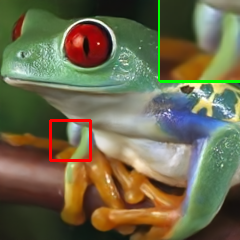} & 
         \includegraphics[trim={0 2cm 0 0},clip, width=0.22\linewidth]{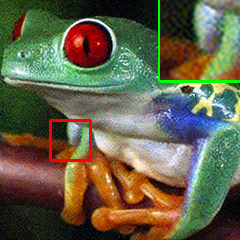} & 
         \includegraphics[trim={0 2cm 0 0},clip, width=0.22\linewidth]{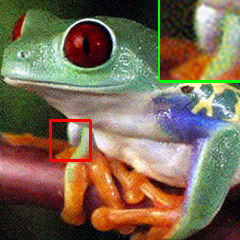} \\
         Input (RealSRSet) & \we & \instp~\#1 & \instp~\#2
    \end{tabular}
    \vspace{-1mm}
    \caption{\textbf{ Comparison with \instp~\cite{brooks2023instructpix2pix}} using the prompt \emph{``Remove the noise in this photo"}. Real-case image from \emph{RealSRSet}~\cite{liang2021swinir}.
    }
    \label{fig:comp-instpix}
\end{figure*}
%%%%%%%%%%%%%%%%%%%%%%%%%%%%%%%%%%%%%%%%%%%%%%%%%%%%%%%%

%%%%%%%%%%%%%%%%%%%%%%%%%%%%%%%%%%%%%%%%%%%%%%%%%%%%%%
%%%%%%%%%%%%%%%%%%%%%%%%%%%%%%%%%%%%%%%%%%%%%%%%%%%%%%

\subsection{On the Effectiveness of Instructions}

Thanks to our integration of human instructions, users can control how to enhance the images. We provide examples in Figures~\ref{fig:comp-instr} and~\ref{fig:multi-instr}, where we show the potential of our method to restore and enhance images in a controllable manner.

%%%%%%%%%%%%% ENDING %%%%%%%%%%%%%%%%%%%%%
%%%%%%%%%%%%%%%%%%%%%%%%%%%%%%%%%%%%%%%%%%

This implies an advancement \emph{w.r.t} classical (deterministic) image restoration methods. Classical deep restoration methods lead to a unique result, thus, they do not allow to control how the image is processed.  We also compare \we with \instp~\cite{brooks2023instructpix2pix} (a diffusion-based generative model) in Figure~\ref{fig:comp-instpix}. %We provide more samples in the supplementary material.

\vspace{-3mm}
\paragraph{Qualitative Results.} 

\begin{figure*}[t]
    \centering
    \setlength{\tabcolsep}{1pt}
    \begin{tabular}{c c c c c}
         \includegraphics[width=0.19\linewidth]{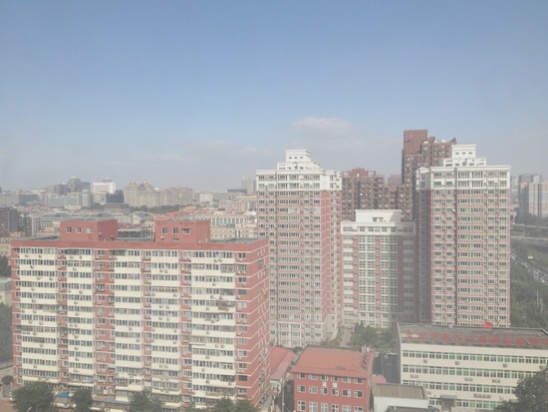} &  
         \includegraphics[width=0.19\linewidth]{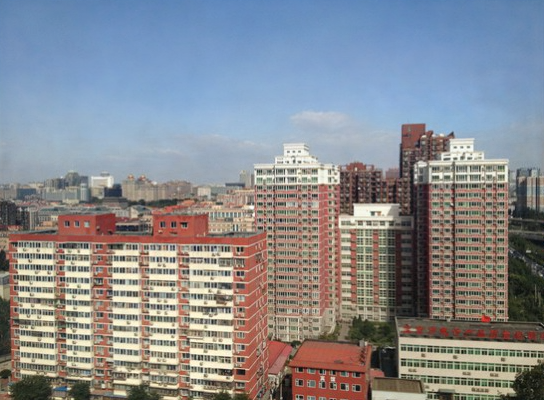} &
         \includegraphics[width=0.19\linewidth]{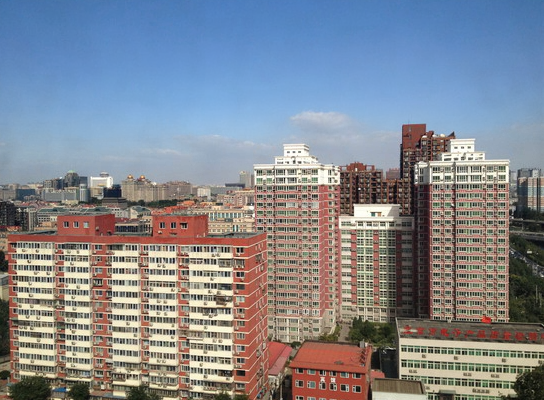} &
         \includegraphics[width=0.19\linewidth]{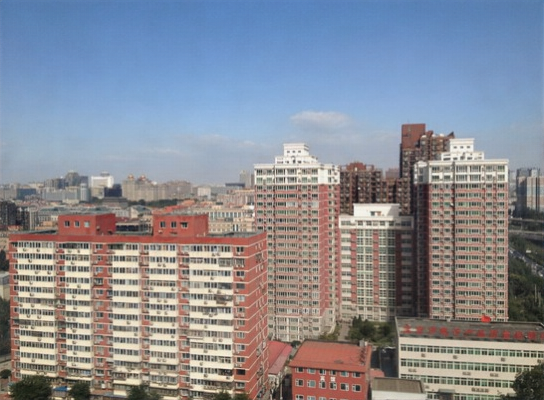} &
         \includegraphics[width=0.19\linewidth]{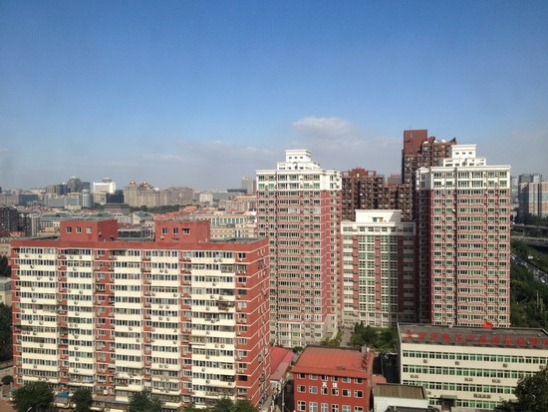} \\
         \includegraphics[width=0.19\linewidth]{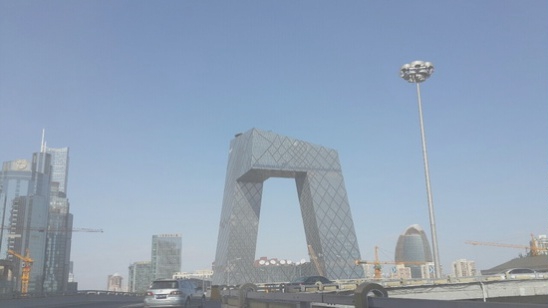} &  
         \includegraphics[width=0.19\linewidth]{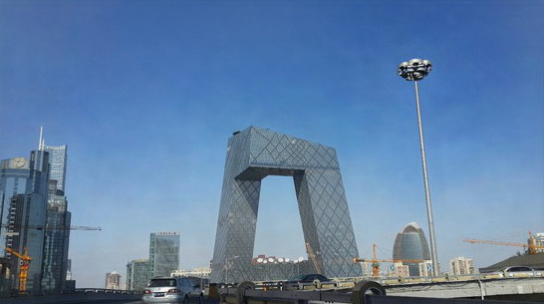} &
         \includegraphics[width=0.19\linewidth]{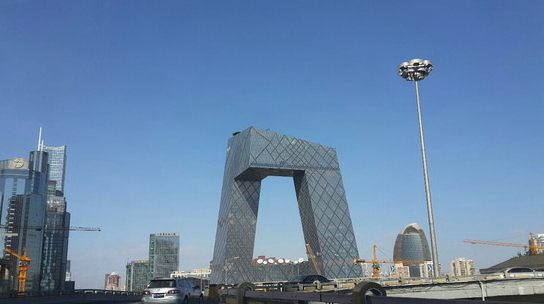} &
         \includegraphics[width=0.19\linewidth]{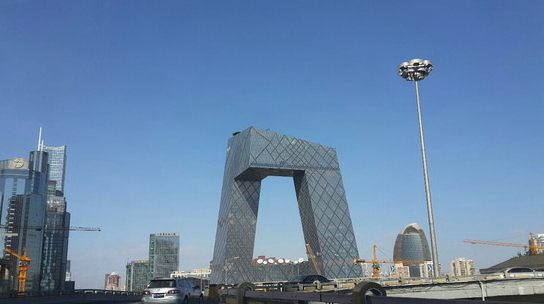} &
         \includegraphics[width=0.19\linewidth]{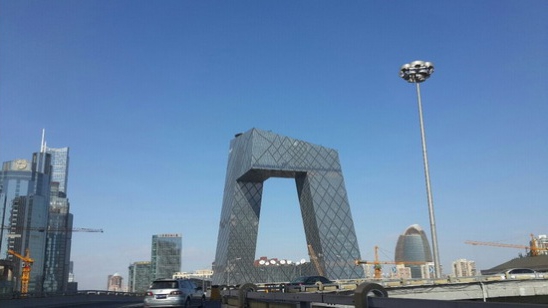} \\
         Input & AirNet~\cite{Li_2022_CVPR} & PromptIR~\cite{potlapalli2023promptir} & \we & Reference \\
    \end{tabular}
    \vspace{-2mm}
    \caption{\textbf{Dehazing comparisons} for all-in-one restoration methods on SOTS~\cite{li2018benchmarking}.
    }
    \label{fig:haze-quali}
\end{figure*}

\begin{figure*}[t]
    \centering
    \includegraphics[width=0.96\linewidth]{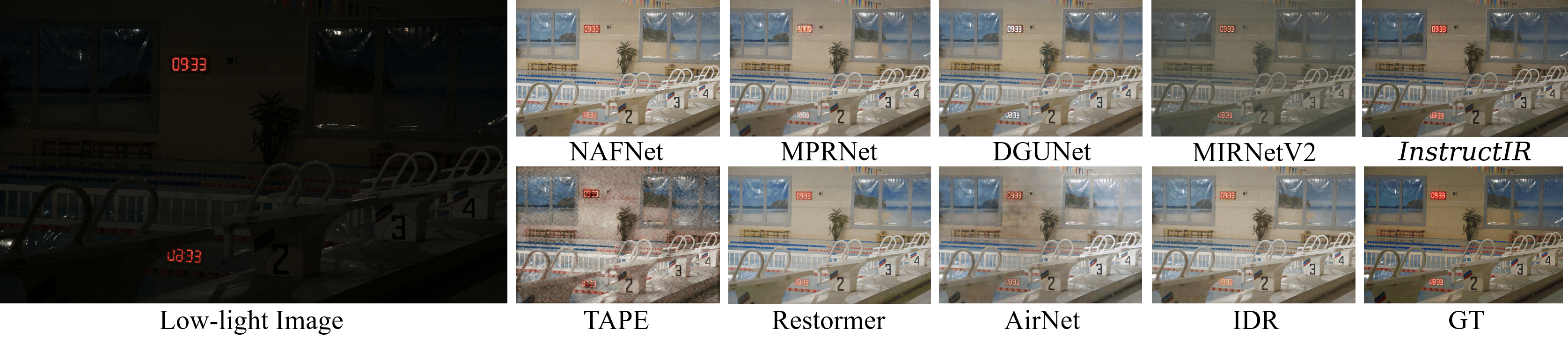}
    \vspace{-2mm}
    \caption{
    \textbf{Real Low-light Enhancement Results.} LOL~\cite{Chen2018Retinex} testset (\texttt{748.png}). AirNet~\cite{Li_2022_CVPR} and IDR~\cite{zhang2023ingredient} are well-known all-in-one restoration methods. NAFNet~\cite{chen2022simple} is equivalent to \we without text conditions (\ie our image-only backbone).}
    \label{fig:lol-comp}
\end{figure*}

\begin{figure*}[t]
    \centering
    \includegraphics[width=0.96\linewidth]{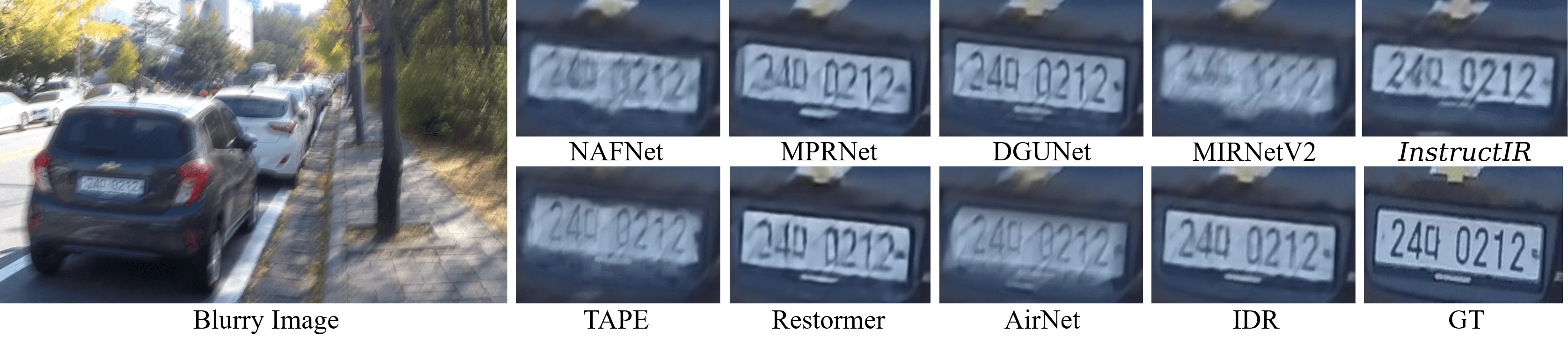}
    \vspace{-4mm}
    \caption{\textbf{Image Deblurring Results.} GoPro~\cite{gopro2017} dataset.}
    %(\texttt{GOPR0854-11-00-000001.png})
    \label{fig:blur-comp}
\end{figure*}

\begin{figure*}[t]
    \centering
    \includegraphics[width=0.96\linewidth]{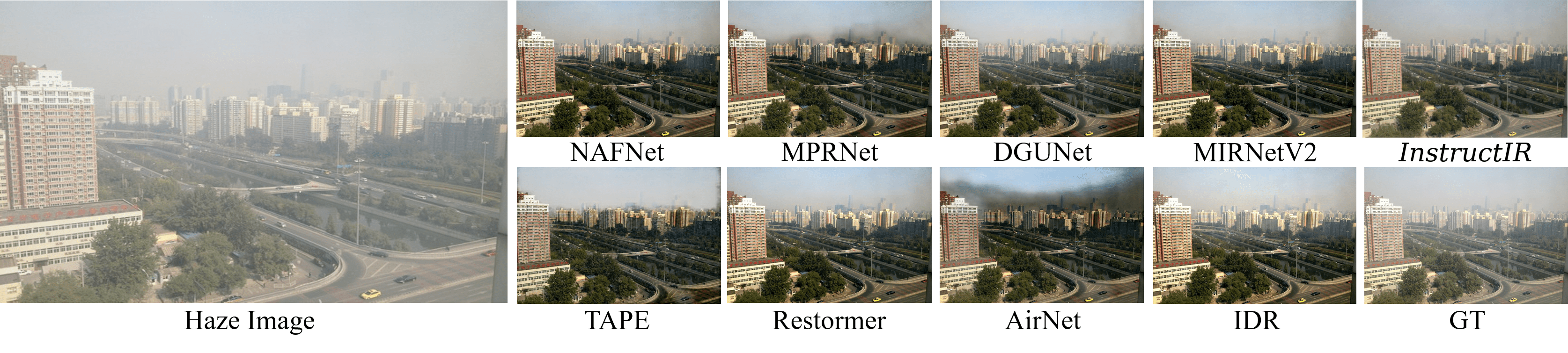}
    \vspace{-4mm}
    \caption{\textbf{Image Dehazing Results.} SOTS~\cite{li2018benchmarking} \emph{outdoor} dataset (\texttt{0150.jpg}).}
    \label{fig:haze-comp}
\end{figure*}

\begin{figure*}[t]
    \centering
    \includegraphics[width=0.96\linewidth]{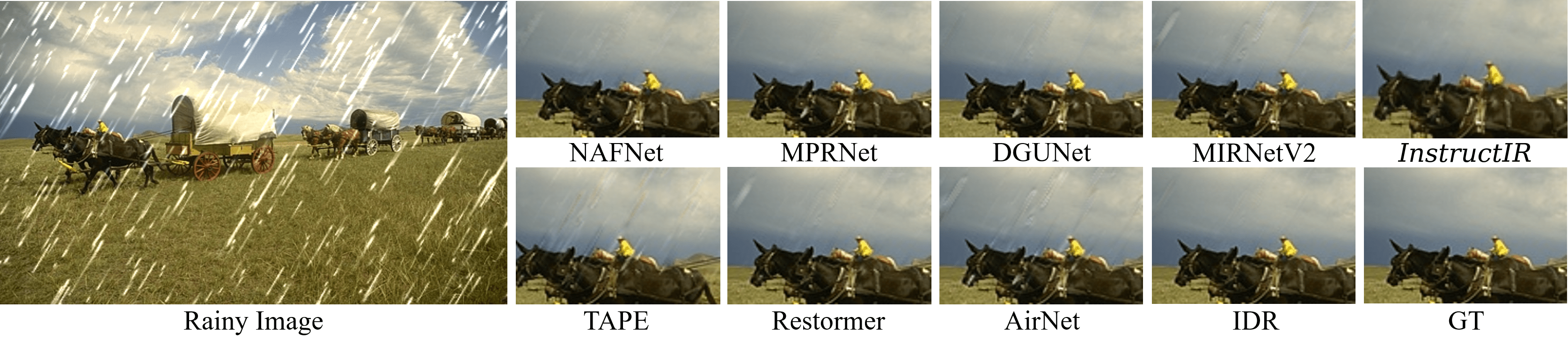}
    \vspace{-4mm}
    \caption{\textbf{Image Deraining Results} on Rain100L~\cite{fan2019general} (\texttt{035.png}).}
    \label{fig:rain-comp}
\end{figure*}

\begin{figure*}[t]
    \centering
    \includegraphics[width=0.96\linewidth]{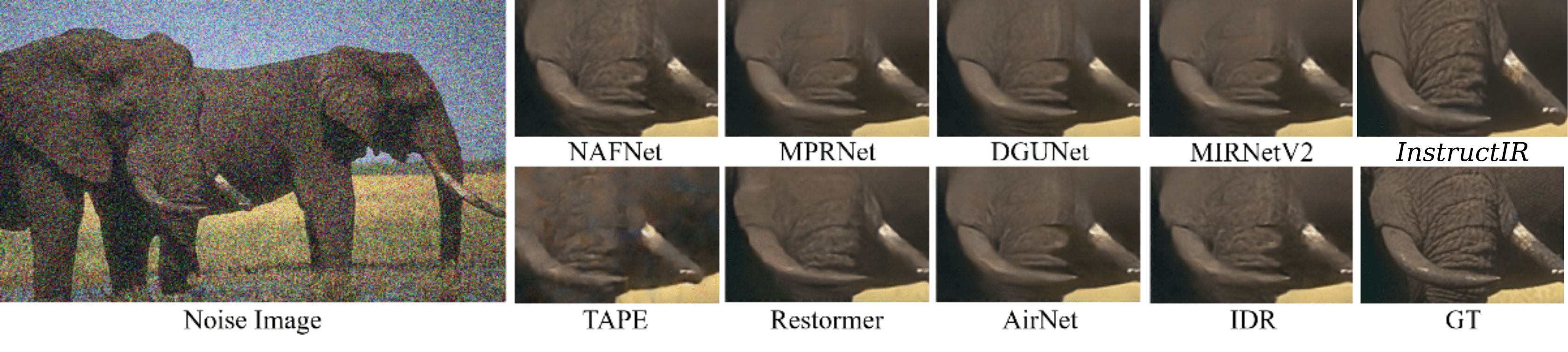}
    \vspace{-4mm}
    \caption{\textbf{Image Denoising Results} on BSD68~\cite{martin2001database_bsd} (\texttt{0060.png}).}
    \label{fig:noise-comp}
\end{figure*}

We provide diverse qualitative results for several tasks, and we compare with all-in-one and task-specific methods. 
In Figure~\ref{fig:lol-comp}, we show results on the LOL~\cite{Chen2018Retinex} dataset. 
In Figure~\ref{fig:blur-comp}, we compare methods on the motion deblurring task using the GoPro~\cite{gopro2017} dataset.
In Figures~\ref{fig:haze-quali}~and~\ref{fig:haze-comp}, we compare with different methods for the dehazing task on SOTS (outdoor)~\cite{li2018benchmarking}. 
In Figure~\ref{fig:rain-comp}, we compare with image restoration methods for deraining on Rain100L~\cite{fan2019general}. Finally, we show denoising results in Figure~\ref{fig:noise-comp}. 
In this qualitative analysis, we use our single \we-\fived model to restore all the images.

%%%%%%%%%%%%%%%%%%%%%%%%%%%%%%%%%%%%%%%%%%%%%%%%

%\vspace{-3mm}
%\paragraph{Discussion on Instruction-based Restoration} In Figure~\ref{fig:comp-instpix2} we compare with \instp~\cite{brooks2023instructpix2pix}. Our method is notably superior in terms of efficiency, fidelity and quality. We can conclude that diffusion-based methods~\cite{rombach2022stablediffusion, meng2021sdedit, brooks2023instructpix2pix} for image manipulation require complex ``tuning" of several (hyper-)parameters, and strong regularization to enforce fidelity and reduce hallucinations. \instp~\cite{brooks2023instructpix2pix} cannot solve inverse problems directly --although it has a good prior for solving Inpainting--, which indicates that such model require restoration-specific training (or fine-tuning). 

%%%%%%%%%%%%%%%%%%%%%%%%%%%%%%%%%%%%%%%%%%%%%%%%%%%%%%%%%%%%

%%%%%%%%%%%%%%%%%%%%%%%%%%%%%%%%%%%%%%%%%%%%%%%%%%%%%%
%%%% REAL-WORLD EXAMPLES
\begin{figure*}[t]
    \centering
    \setlength{\tabcolsep}{1pt}
    \begin{tabular}{c c c c }
         \rowcolor{lyellow} \multicolumn{4}{c}{\scriptsize{Instruction: \emph{``my colors are too off, make it pop so I can use these photos in instagram''}}} \\
         \includegraphics[trim={0 0 0 0},clip, width=0.23\linewidth]{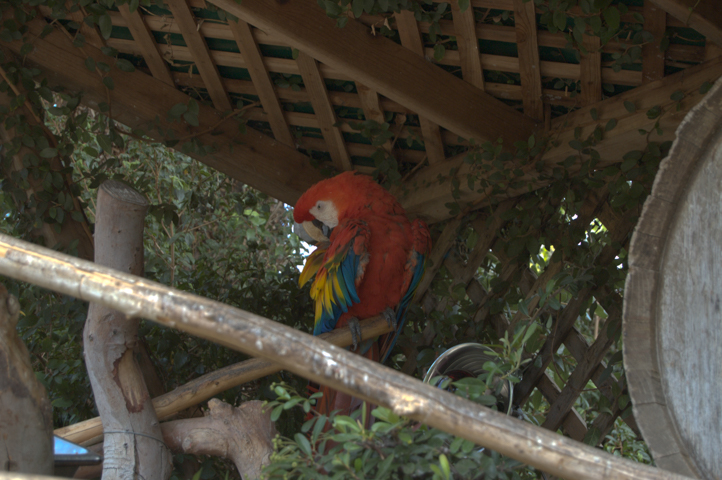} & 
         \includegraphics[trim={0 0 0 0},clip, width=0.23\linewidth]{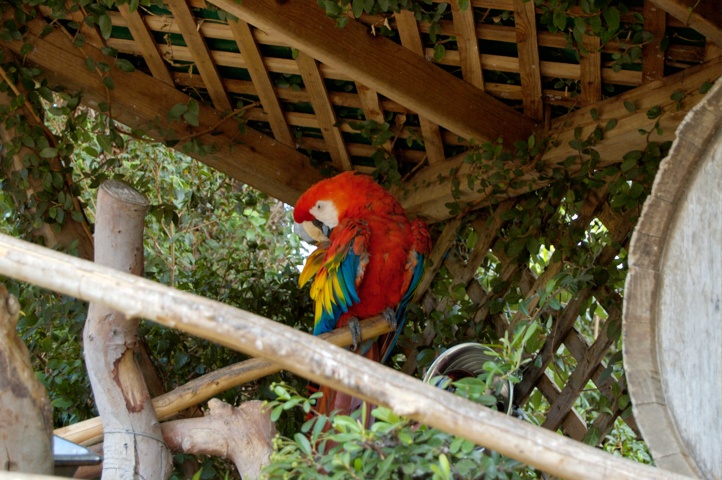} & 
         \includegraphics[trim={0 0 0 0},clip, width=0.23\linewidth]{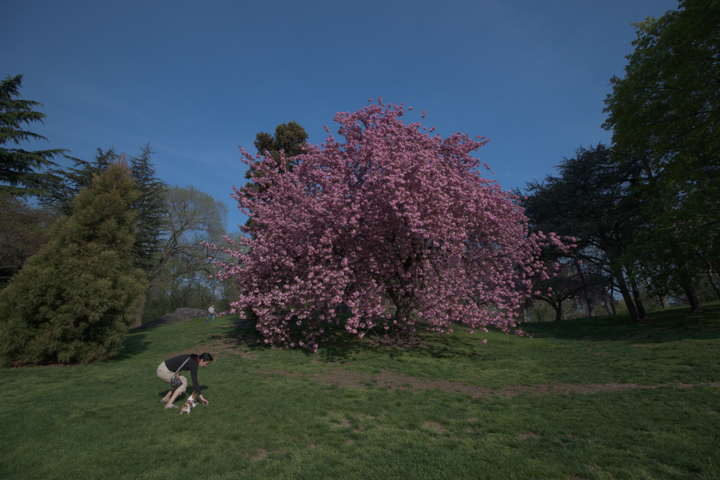} & 
         \includegraphics[trim={0 0 0 0},clip, width=0.23\linewidth]{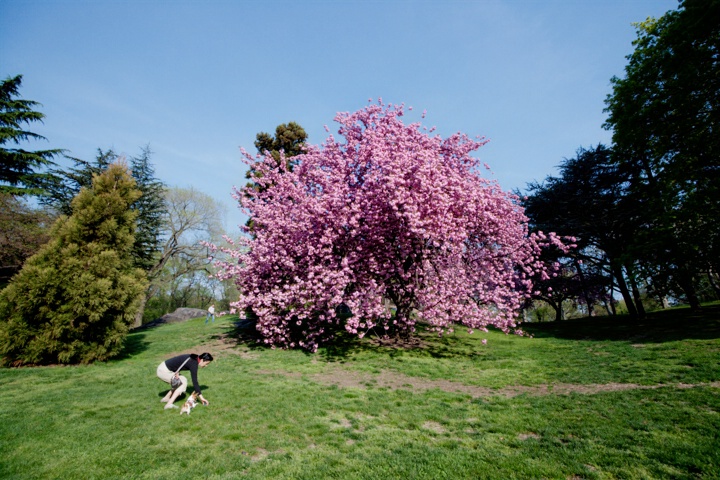} \\
         %%%
         \rowcolor{lyellow} \multicolumn{4}{c}{\scriptsize{\emph{``Increase the brightness of my photo please, is it totoro?''~~|~~``my image is too dark, fix it''}}} \\
         \includegraphics[trim={0 0 0 0},clip, width=0.23\linewidth]{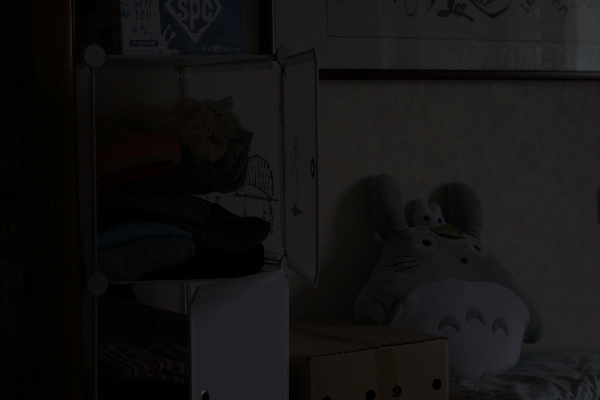} & 
         \includegraphics[trim={0 0 0 0},clip, width=0.23\linewidth]{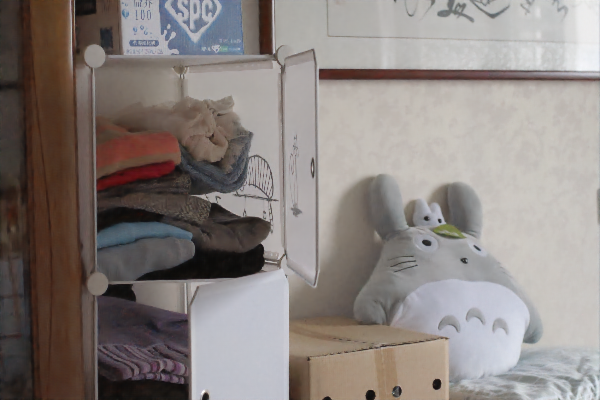} & 
         \includegraphics[trim={0 0 0 0},clip, width=0.23\linewidth]{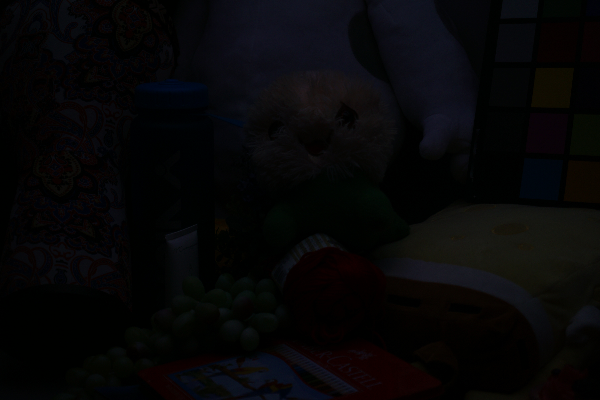} & 
         \includegraphics[trim={0 0 0 0},clip, width=0.23\linewidth]{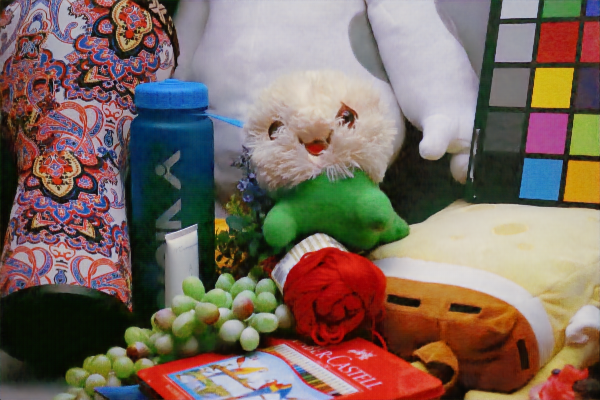} \\
    \end{tabular}
    \vspace{-1mm}
    \caption{\textbf{Real-world} samples of image restoration and enhancement using \we.
    }
    \label{fig:real-results}
\end{figure*}
%%%%%%%%%%%%%%%%%%%%%%%%%%%%%%%%%%%%%%%%%%%%%%%%%%%%%%%%
%%%% ALL-IN-ONE DEHAZE
%%%%%%%%%%%%%%%%%%%%%%%%%%%%%%%%%%%%%%%%%%%%%%%%%%%%%%%%%%%

\vspace{-3mm}
\paragraph{Limitations}
%Our method achieves \emph{state-of-the-art} results in five tasks, proving the potential of using instructions to guide deep blind restoration models. However, we acknowledge certain limitations. First, in comparison to diffusion-based restoration methods, our current approach would not produce better results attending to perceptual quality. 
As with previous \emph{all-in-one} methods, our model struggles to process images with more than one degradation (\ie complex \emph{real-world} images), or unknown out-of-distribution degradations, yet this is a common limitation among the related methods. However, we believe that these limitations can be surpassed with more realistic training data, and scaling the model's complexity.

%%%%%%%%%%%%%%%%%%%%%%%%%%%%%%%%%%%%%%%%%%%%%%%%%%%%%%%%%%%
\vspace{-2mm}
\section{Conclusion}
We present a novel approach that uses natural human-written instructions to guide the image restoration model. Given a prompt, our multi-task model can recover high-quality images from their degraded counterparts, considering multiple degradations. We achieve state-of-the-art results on several restoration tasks, demonstrating the power and flexibility of instruction guidance. Our results represent a new benchmark for text-guided image restoration.

\clearpage

%%%%%%%%%%%%%%%%%%%%%%%%%%%%%%%%%%%%%%%%%%%%%%%%%%%
%%%%%%%%%%%%%%%%%%%%%%%%%%%%%%%%%%%%%%%%%%%%%%%%%%%
%%% LOL
%%%%%%%%%%%%%%%%%%%%%%%%%%%%%%%%%%%%%%%%%%%%%%%%%%%
%%% GOPRO
\vspace{-15mm}
%%%%%%%%%%%%%%%%%%%%%%%%%%%%%%%%%%%%%%%%%%%%%%%%%%%
%%% HAZE
\vspace{-15mm}
%%%%%%%%%%%%%%%%%%%%%%%%%%%%%%%%%%%%%%%%%%%%%%%%%%%
%%% RAIN
\vspace{-15mm}
%%%%%%%%%%%%%%%%%%%%%%%%%%%%%%%%%%%%%%%%%%%%%%%%%%%
%%% NOISE
\vspace{-15mm}
%%%%%%%%%%%%%%%%%%%%%%%%%%%%%%%%%%%%%%%%%%%%%%%%%%%%%%


\clearpage

\subsection*{Acknowledgments} 
This work was partly supported by the The Humboldt Foundation (AvH). Work done at the University of Würzburg. Marcos Conde is also supported by Sony Interactive Entertainment, FTG.
%}

%%%%%%%%%%%%%%%%%%%%%%%%%%%%%%%%%%%%%%%%%%%%%%%%%
\appendix

\section{Additional Training Details and Ablations}

\begin{figure}[!ht]
    \centering
    \includegraphics[width=0.7\linewidth]{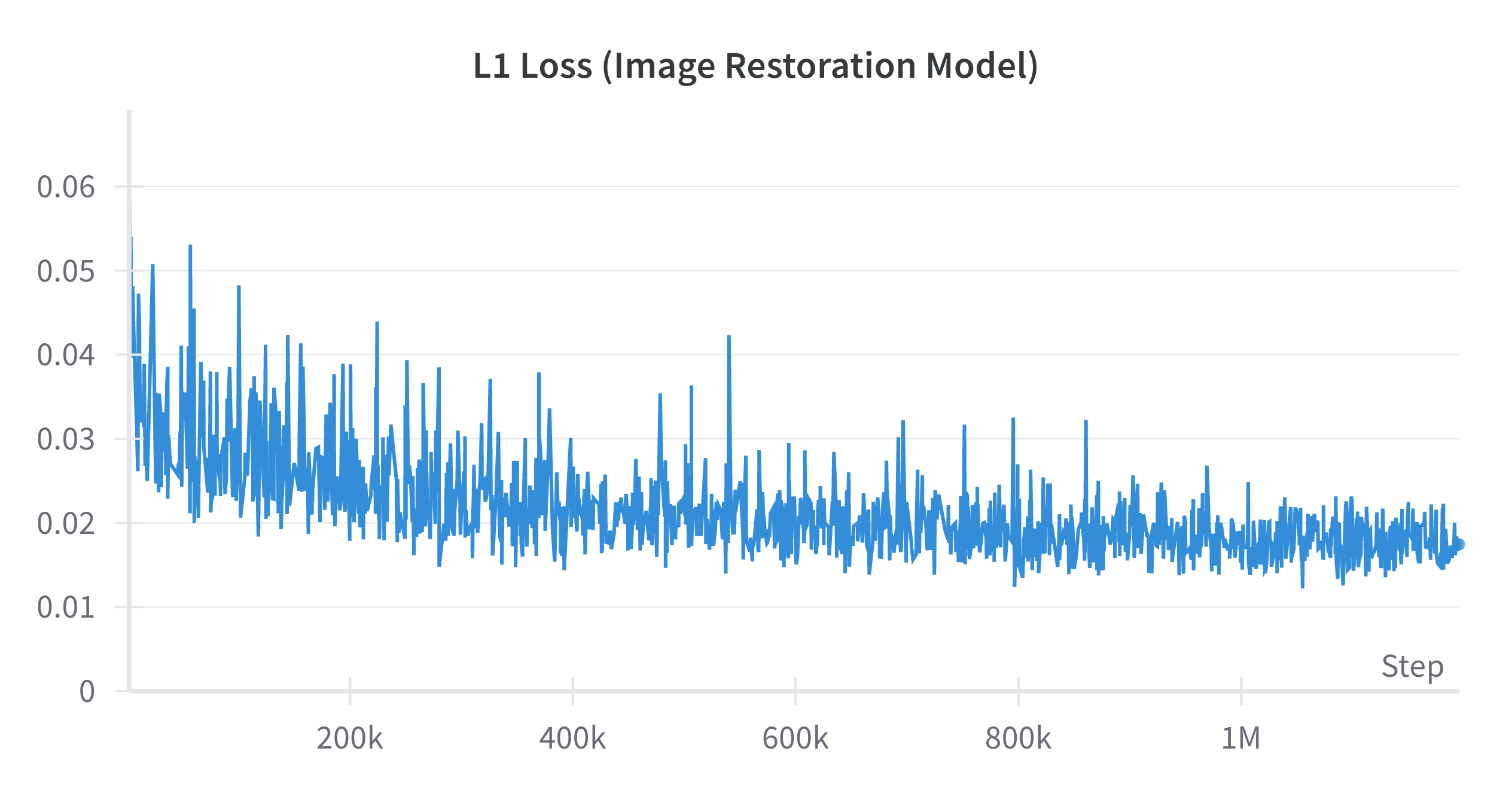}
    \caption{Image Restoration Loss ($\mathcal{L}_1$) computed between the restored image $\hat{x}$ (model's output) and the reference image $x$.}
    \label{fig:ir-loss}
\end{figure}

\begin{figure}[!ht]
    \centering
    \includegraphics[width=0.7\linewidth]{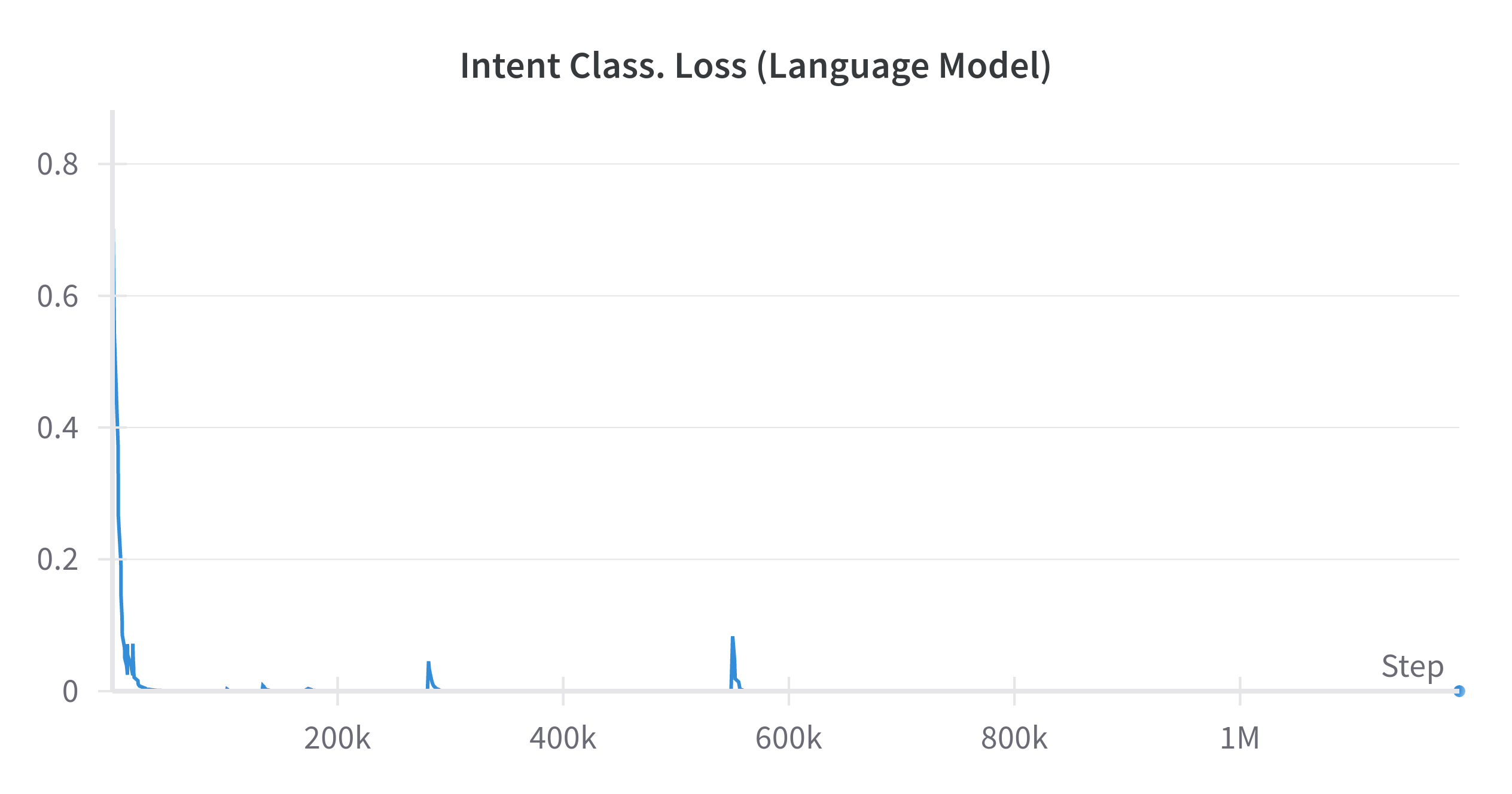}
    \caption{Intent Classification Loss from the instructions. Product of our simple MLP classification head using $\mathbf{e}$. When $\mathcal{L}_{ce}\!\to\!0$ the model uses the learned prompt embeddings, and it is optimized mainly using the image regression loss ($\mathcal{L}_1$).
    }
    \label{fig:lm-loss}
\end{figure}

We define our loss functions in the paper \emph{Sec.~4.1}. Our training loss function is $\mathcal{L} = \mathcal{L}_1 + \mathcal{L}_{ce}$, which includes the loss function of the image model ($\mathcal{L}_1$), and the loss function for intent (task/degradation) classification ($\mathcal{L}_{ce}$) given the prompt embedding. We provide the loss evolution plots in Figures~\ref{fig:ir-loss} and~\ref{fig:lm-loss}. In particular, in Figure~\ref{fig:lm-loss} we can observe how the intent classification loss (\ie predicting the task (or degradation) given the prompt), tends to 0 very fast, indicating that our language model component can infer easily the task given the instruction.

\vspace{2mm}
\noindent Additionally, we study three different text (sentence) encoders: (i) \textsc{BGE-micro-v2}~\footnote{\url{https://huggingface.co/TaylorAI/bge-micro-v2}}, (ii) \textsc{all-MiniLM-L6-v2}~\footnote{\url{https://huggingface.co/sentence-transformers/all-MiniLM-L6-v2}}, (iii) CLIP text encoder (OpenAI CLIP ViT B-16). Note that these are always frozen. We use pre-trained weights from HuggingFace.

In Table~\ref{tab:ab-encoder} we show the ablation study. There is no significant difference between the text encoders. This is related to the previous results (Fig.~\ref{fig:lm-loss}), any text encoder with enough complexity can infer the task from the prompt. Therefore, we use \textsc{BGE-micro-v2}, as it is just 17M parameters in comparison to the others (40-60M parameters). \emph{Note that for this ablation study, we keep \underline{fixed} the image model (16M), and we only change the language model.}

\vspace{2mm}
\paragraph{Text Discussion} We shall ask, \emph{do the text encoders perform great because the language and instructions are too simple?}

We believe our instructions cover a wide range of expressions (technical, common language, ambiguous, etc). The language model works properly on real-world instructions. Therefore, we believe the language for this specific task is self-constrained, and easier to understand and to model in comparison to other ``open" tasks such as image generation.

\vspace{2mm}
\paragraph{Model Design} Based on our experiments, given a trained text-guided image model (\eg based on NAFNet~\cite{chen2022simple}), we can switch language models without performance loss.

%%%%%%%%%%%%%%%%%%%%%%%%%%%%%%%%%%%%%%%%%%%%%%%
\begin{table}[!ht]
    \centering
    \caption{\textbf{Ablation study on the text encoders.} We report PSNR/SSIM metrics for each task using our \textcolor{purple}{\bf 5D} base model. We use the same \underline{fixed} image model (based on NAFNet~\cite{chen2022simple}).}
    \label{tab:ab-encoder}
    \resizebox{0.65\linewidth}{!}{
    \begin{tabular}{l c c c c}
    \toprule
    \textbf{Encoder}    & \textbf{Deraining} & \textbf{Denoising} & \textbf{Deblurring} & \textbf{LOL}  \\
    \midrule
    \rowcolor{lyellow} \textsc{BGE-micro}   & 36.84/0.973 & 31.40/0.887 & 29.40/0.886  & 23.00/0.836 \\
    \textsc{ALL-MINILM}  & 36.82/0.972 & 31.39/0.887 & 29.40/0.886  & 22.98/0.836 \\
    \textsc{CLIP}  & 36.83/0.973 & 31.39/0.887 & 29.40/0.886  & 22.95/0.834 \\
    \bottomrule
    \end{tabular}
    }
\end{table}
%%%%%%%%%%%%%%%%%%%%%%%%%%%%%%%%%%%%%%%%%%%%%%%
%%%%%%%%%%%%%%%%%%%%%%%%%%%%%%%%%%%%%%%%%%%%%%%
%%%%%%%%%%%%%%%%%%%%%%%%%%%%%%%%%%%%%%%%%%%%%%%

\vspace{1mm}
\noindent\emph{Comparison of NAFNet with and without using text (i.e. image only)}: The reader can find the comparison in the main paper Table 2, please read the highlighted caption.

\vspace{1mm}
\noindent\emph{How the 6D variant does Super-Resolution?}. We degraded the input images by downsampling and re-upsampling using Bicubic interpolation. Given a LR image, we updample it using Bicubic, then InstructIR can recover some details. As we discuss in the paper, adding this task helps the main task of deblurring.

%\vspace{1mm}
%\paragraph{Real-World Generalization.} We evaluate \we as previous works~\cite{potlapalli2023promptir, Li_2022_CVPR, zhang2023ingredient}. Also, we find the same limitations as such methods when we process real-world images. Evaluating the model on (multiple) real-world degradations is a future task.

%%%%%%%%%%%%%%%%%%%%%%%%%%%%%%%%%%%%%%%%%%%%%%%

\vspace{-2mm}
\paragraph{Contemporary Works and Reproducibility.} Note that PromptIR, ProRes~\cite{ma2023prores} and Amirnet~\cite{zhang2023allamirnet} are contemporary works (presented or published by Dec 2023). We compare mainly with AirNet~\cite{Li_2022_CVPR} since the model and results are open-source, and it is a reference all-in-one method. To the best of our knowledge, IDR~\cite{zhang2023ingredient} and ADMS~\cite{park2023all} do not provide open-source code, models or results, thus we cannot compare with them qualitatively.

\vspace{-2mm}
\subsection{Additional Ablation Studies}

We provide ablation studies and comparison with more task-specific methods in Tables~\ref{tab:noise} (image denoising) and Table~\ref{tab:gopro-sots} (image deblurring and dehazing).

\begin{table}[!ht]
    \centering
    \caption{Comparison with general restoration and all-in-one methods (\redast) at \textbf{image denoising}. We report PSNR on benchmark datasets considering different $\sigma$ noise levels. Table based on~\cite{zhang2023ingredient}.}
    \vspace{-1mm}
    \label{tab:noise}
    \resizebox{0.8\linewidth}{!}{
    \setlength{\tabcolsep}{3pt}
    \begin{tabular}{r | ccc | ccc | ccc}
         \toprule
         &      \multicolumn{3}{c|}{\bf CBSD68~\cite{martin2001database_bsd}} & \multicolumn{3}{c|}{\bf Urban100~\cite{huang2015single}} & \multicolumn{3}{c}{\bf Kodak24~\cite{kodak}}\\
         \cmidrule(lr){2-4} \cmidrule(lr){5-7} \cmidrule(lr){8-10}
         \textbf{Method} & \textbf{15} & \textbf{25} & \textbf{50} & \textbf{15} & \textbf{25} & \textbf{50} & \textbf{15} & \textbf{25} & \textbf{50} \\
         \midrule
         IRCNN~\cite{zhang2017learning}   & 33.86 & 31.16 & 27.86 & 33.78 & 31.20 & 27.70 & 34.69 & 32.18 & 28.93 \\

         FFDNet~\cite{FFDNetPlus}  & 33.87 & 31.21 & 27.96 & 33.83 & 31.40 & 28.05 & 34.63 & 32.13 & 28.98 \\
         
         DnCNN~\cite{DnCNN}  & 33.90 & 31.24 & 27.95 & 32.98 & 30.81 & 27.59 & 34.60 & 32.14 & 28.95 \\
         NAFNet~\cite{chen2022simple}     & 33.67 & 31.02 & 27.73 & 33.14 & 30.64 & 27.20 & 34.27 & 31.80 & 28.62 \\
         HINet~\cite{chen2021hinet}      & 33.72 & 31.00 & 27.63 & 33.49 & 30.94 & 27.32 & 34.38 & 31.84 & 28.52 \\
         DGUNet~\cite{mou2022deep}     & 33.85 & 31.10 & 27.92 & 33.67 & 31.27 & 27.94 & 34.56 & 32.10 & 28.91 \\
         MIRNetV2~\cite{zamir2020mirnet}   & 33.66 & 30.97 & 27.66 & 33.30 & 30.75 & 27.22 & 34.29 & 31.81 & 28.55 \\
         SwinIR~\cite{liang2021swinir}     & 33.31 & 30.59 & 27.13 & 32.79 & 30.18 & 26.52 & 33.89 & 31.32 & 27.93 \\
         Restormer~\cite{restormer}  & 34.03 & 31.49 & 28.11 & 33.72 & 31.26 & 28.03 & 34.78 & 32.37 & 29.08 \\
         \midrule
         \midrule
         \redast DL~\cite{fan2019general}             & 23.16 & 23.09 & 22.09 & 21.10 & 21.28 & 20.42 & 22.63 & 22.66 & 21.95 \\
         \redast T.weather~\cite{valanarasu2022transweather}   & 31.16 & 29.00 & 26.08 & 29.64 & 27.97 & 26.08 & 31.67 & 29.64 & 26.74 \\
         \redast TAPE~\cite{liu2022tape}           & 32.86 & 30.18 & 26.63 & 32.19 & 29.65 & 25.87 & 33.24 & 30.70 & 27.19 \\
         \redast AirNet~\cite{Li_2022_CVPR}         & 33.49 & 30.91 & 27.66 & 33.16 & 30.83 & 27.45 & 34.14 & 31.74 & 28.59 \\
         \redast IDR~\cite{zhang2023ingredient}            & 34.11 & 31.60 & 28.14 & 33.82 & 31.29 & 28.07 & 34.78 & 32.42 & 29.13 \\
         \rowcolor{lyellow} \redast \we-\fived & 
         34.00 & 31.40 & 28.15 & 
         33.77 & 31.40 & 28.13 & 
         34.70 & 32.26 & 29.16 \\
         \rowcolor{lyellow} \redast \we-\threed & 
         \textbf{34.15} & \textbf{31.52} & \textbf{28.30} & 
         \textbf{34.12} & \textbf{31.80} & \textbf{28.63} & 
         \textbf{34.92} & \textbf{32.50} & \textbf{29.40} \\
         \bottomrule
    \end{tabular}
    }
\end{table}

\begin{table}[!ht]
    \centering
    \caption{\textbf{Deblurring and Dehazing comparisons.} We compare with task-specific classical methods on benchmark datasets.}
    \vspace{-1mm}
    \label{tab:gopro-sots}
    \resizebox{0.7\linewidth}{!}{
    \begin{tabular}{r c || r c}
         \toprule
         \multicolumn{2}{c||}{\bf Deblurring GoPro~\cite{gopro2017}} & \multicolumn{2}{c}{\bf Dehazing SOTS~\cite{li2018benchmarking}}\\
         \rule{0pt}{3ex}
         \textbf{Method} & \textbf{PSNR/SSIM} & \textbf{Method} & \textbf{PSNR/SSIM} \\
         \midrule
         Xu~\emph{et al.}~\cite{xu2013unnatural}  & 21.00/0.741 & 
         DehazeNet~\cite{cai2016dehazenet} & 22.46/0.851 \\
         
         DeblurGAN~\cite{deblurgan} & 28.70/0.858 & 
         GFN~\cite{ren2018gated-gfn-haze} & 21.55/0.844 \\
         
         Nah~\emph{et al.}~\cite{gopro2017}  & 29.08/0.914 & 
         GCANet~\cite{chen2019gated-gcanet-haze} & 19.98/0.704 \\
         
         RNN~\cite{zhang2018dynamic} & 29.19/0.931 & 
         MSBDN~\cite{dong2020multi-msbdn-haze} & 23.36/0.875 \\
         
         DeblurGAN-v2~\cite{deblurganv2} & \underline{29.55/0.934} 
         & DuRN ~\cite{liu2019dual-durn}     & 24.47/0.839 \\
         \rowcolor{lyellow} \we-\fived & 29.40/0.886 & 
         \we-\fived   & \underline{27.10/0.956} \\
         \rowcolor{lyellow} \we-\sixd  & \textbf{29.73/0.892} & 
         \we-\threed  & \textbf{30.22/0.959} \\
         \bottomrule
    \end{tabular}
    }
\end{table}

\vspace{-2mm}
\section{Additional Visual Results}
\begin{figure*}[!ht]
    \centering
    \setlength{\tabcolsep}{1pt}
    \begin{tabular}{c c c c c}
         \includegraphics[width=0.19\linewidth]{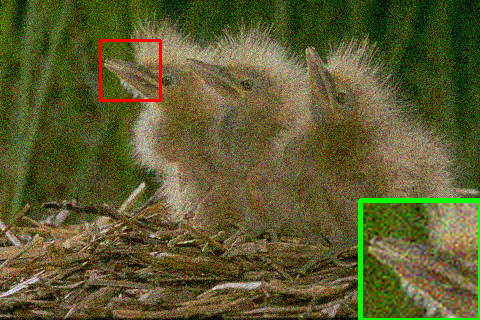} &  
         \includegraphics[width=0.19\linewidth]{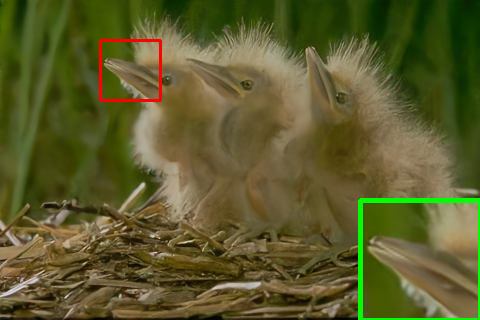} &
         \includegraphics[width=0.19\linewidth]{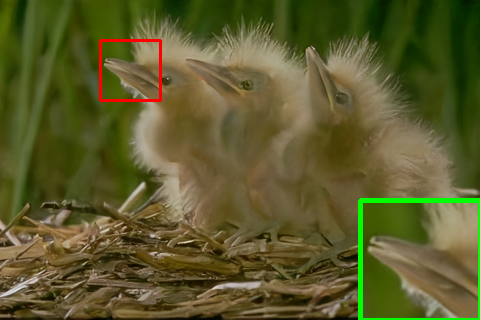} &
         \includegraphics[width=0.19\linewidth]{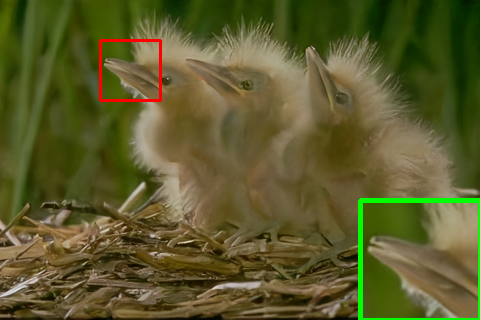} &
         \includegraphics[width=0.19\linewidth]{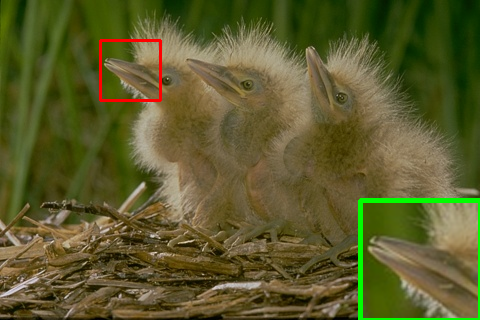} \\
         \includegraphics[width=0.19\linewidth]{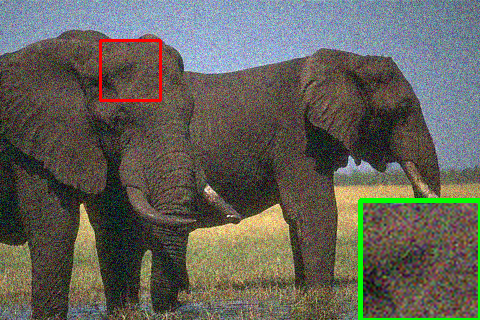} &  
         \includegraphics[width=0.19\linewidth]{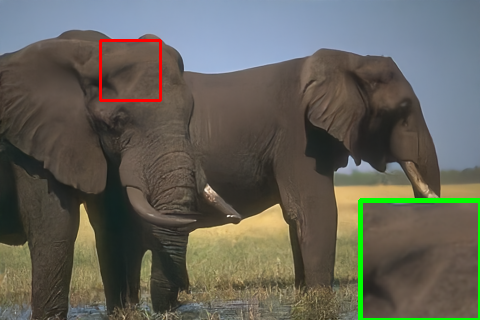} &
         \includegraphics[width=0.19\linewidth]{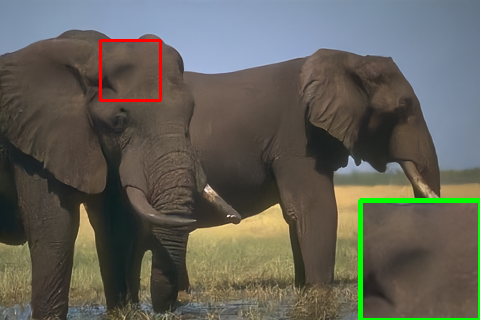} &
         \includegraphics[width=0.19\linewidth]{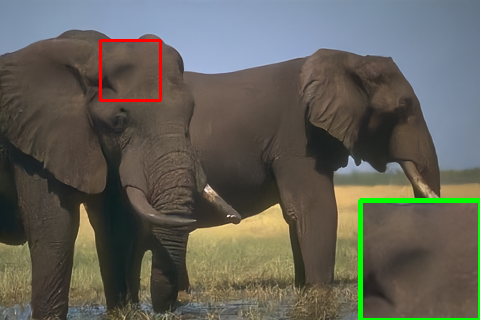} &
         \includegraphics[width=0.19\linewidth]{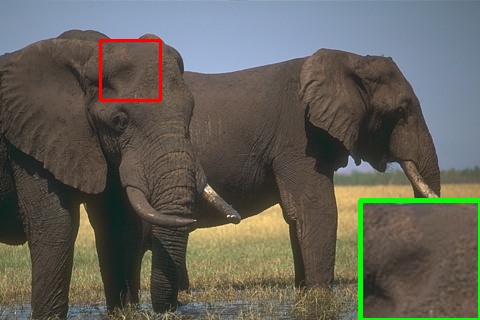} \\
         Input & AirNet~\cite{Li_2022_CVPR} & PromptIR~\cite{potlapalli2023promptir} & \we & Reference \\
    \end{tabular}
    \vspace{-2mm}
    \caption{\textbf{Denoising results} for all-in-one methods. Images from BSD68~\cite{martin2001database_bsd} with noise level $\sigma=25$. }
    \label{fig:noise-quali}
\end{figure*}

\begin{figure*}[!ht]
    \centering
    \setlength{\tabcolsep}{1pt}
    \begin{tabular}{c c c c c}
         \includegraphics[width=0.19\linewidth]{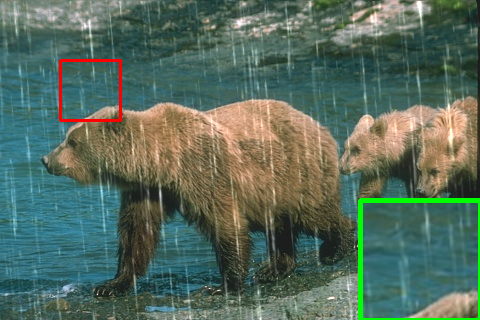} &  
         \includegraphics[width=0.19\linewidth]{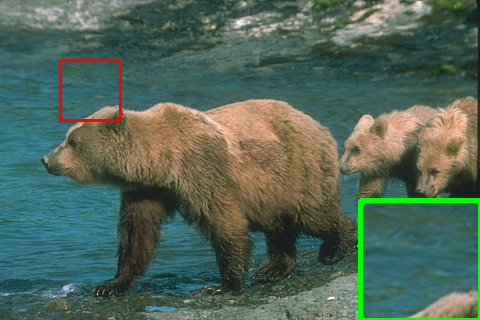} &
         \includegraphics[width=0.19\linewidth]{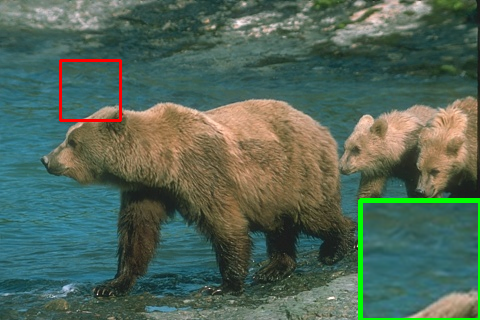} &
         \includegraphics[width=0.19\linewidth]{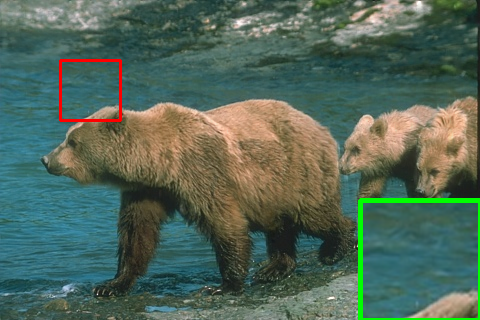} &
         \includegraphics[width=0.19\linewidth]{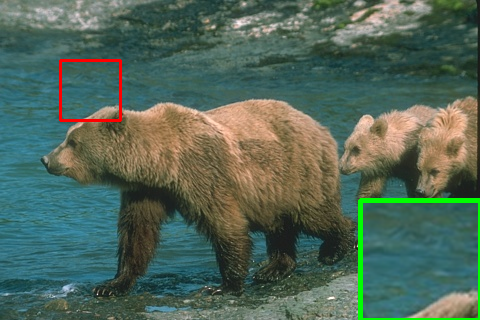} \\
         \includegraphics[width=0.19\linewidth]{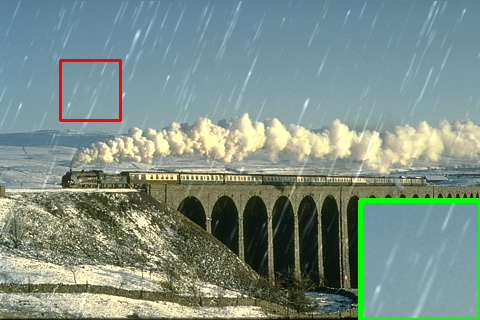} &  
         \includegraphics[width=0.19\linewidth]{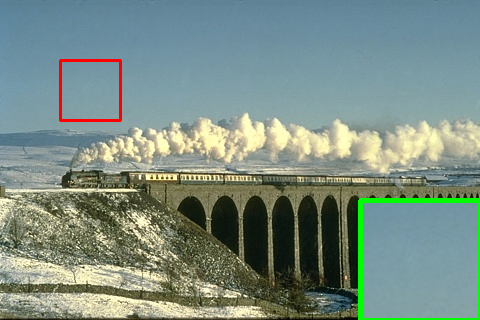} &
         \includegraphics[width=0.19\linewidth]{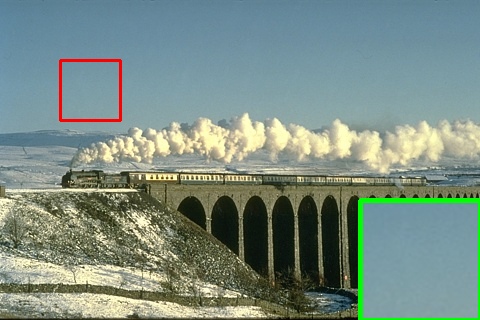} &
         \includegraphics[width=0.19\linewidth]{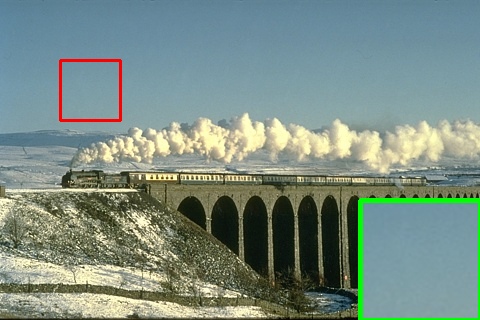} &
         \includegraphics[width=0.19\linewidth]{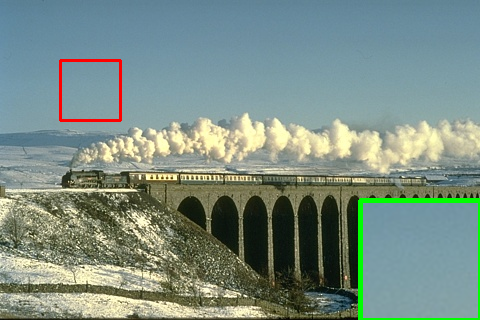} \\
         Input & AirNet~\cite{Li_2022_CVPR} & PromptIR~\cite{potlapalli2023promptir} & \we & Reference \\
    \end{tabular}
    \vspace{-2mm}
    \caption{\textbf{Image deraining comparisons} for all-in-one methods on images from the Rain100L dataset~\cite{fan2019general}. %Our method effectively removes rain streaks to generate rain-free images. 
    }
    \label{fig:rain-quali}
\end{figure*}

\begin{figure*}[!ht]
    \centering
    \setlength{\tabcolsep}{1pt}
    \begin{tabular}{c c c}
         \includegraphics[width=0.32\linewidth]{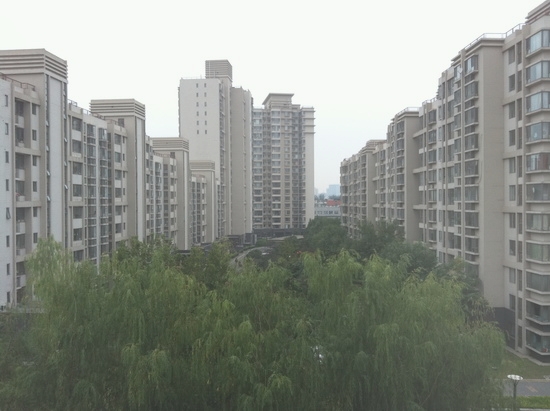} &  
         \includegraphics[width=0.32\linewidth]{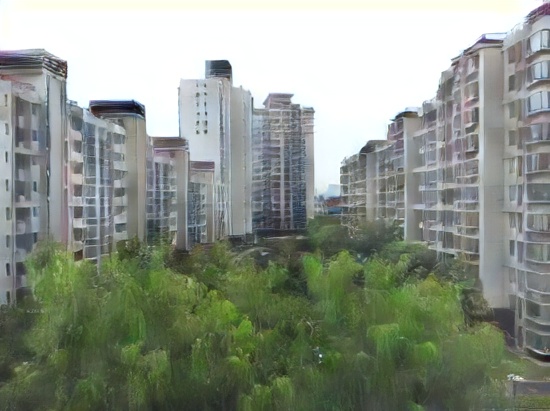} &
         \includegraphics[width=0.32\linewidth]{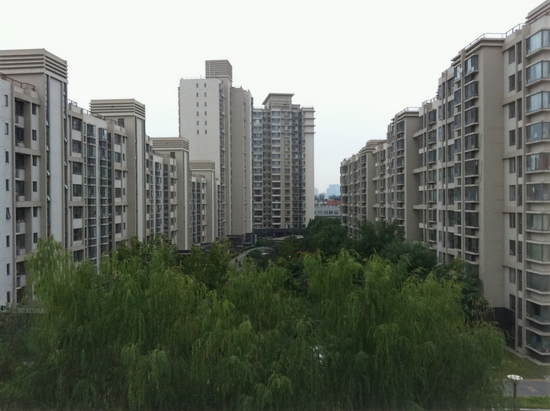} \\
         Input Hazy & RIDCP~\cite{wu2023ridcp} & \we \\
         \includegraphics[width=0.32\linewidth]{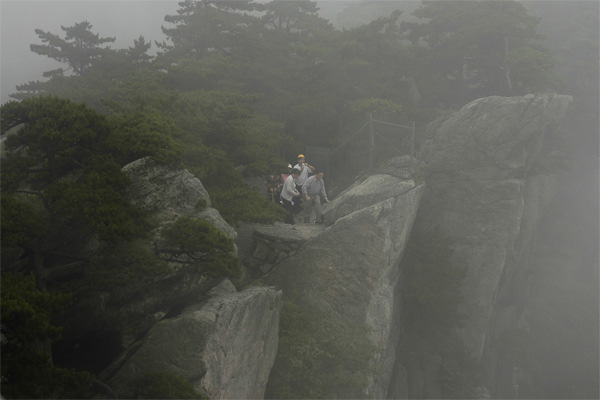} &  
         \includegraphics[width=0.32\linewidth]{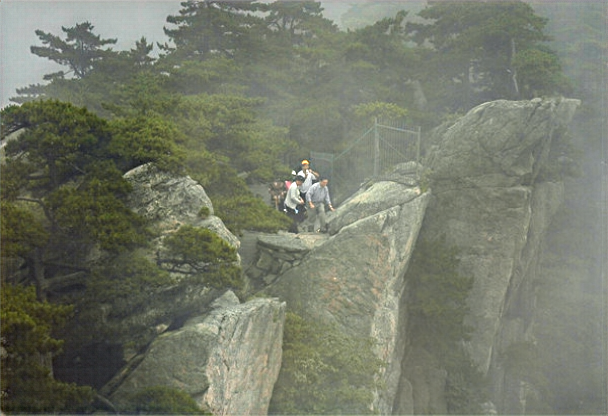} &
         \includegraphics[width=0.32\linewidth]{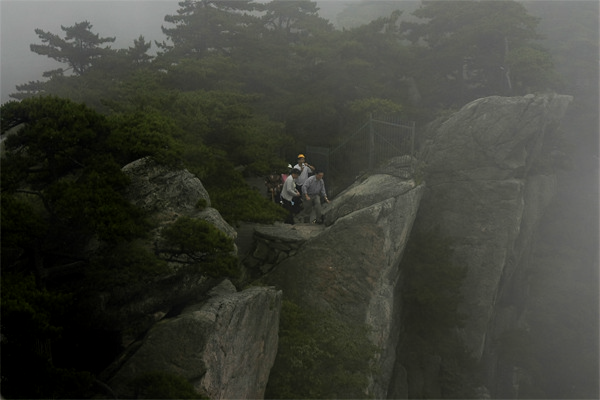} \\
         Input Hazy & PSD~\cite{chen2021psd} & \we \\
    \end{tabular}
    \vspace{-2mm}
    \caption{\textbf{Real Image dehazing comparisons}. These are real-world samples without ground-truth. Our method achieves pleasant results as generative models such as RIDCP~\cite{wu2023ridcp} based on VQGAN. Sample from the RTTS dataset~\cite{li2018benchmarking}. We use the instruction ``remove and haze and mist from this photo please''.
    }
    \label{fig:real-haze-q}
\end{figure*}

\begin{figure*}[!ht]
    \centering
    \setlength{\tabcolsep}{1pt}
    \begin{tabular}{c c c c}
         \midrule
         \rowcolor{lyellow} \multicolumn{4}{c}{Instruction: \emph{``Reduce the noise in this photo"} -- Basic \& Precise}\\
         \midrule
         \includegraphics[trim={0 2cm 0 0},clip, width=0.24\linewidth]{figs/multi-deg-zoom/frog.png} & 
         \includegraphics[trim={0 2cm 0 0},clip, width=0.24\linewidth]{figs/multi-deg-zoom/frog_ours.png} & 
         \includegraphics[trim={0 2cm 0 0},clip, width=0.24\linewidth]{figs/multi-deg-zoom/frog_pix2pix.png} & 
         \includegraphics[trim={0 2cm 0 0},clip, width=0.24\linewidth]{figs/multi-deg-zoom/frog_p2p2.png} \\
         \midrule
         \rowcolor{lyellow} \multicolumn{4}{c}{Instruction: \emph{``Remove the tiny dots in this image"} -- Basic \& Ambiguous }\\
         \midrule
         \includegraphics[trim={0 2cm 0 0},clip, width=0.24\linewidth]{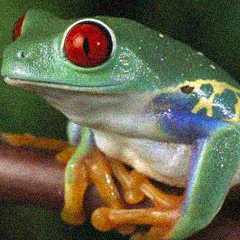} & 
         \includegraphics[trim={0 2cm 0 0},clip, width=0.24\linewidth]{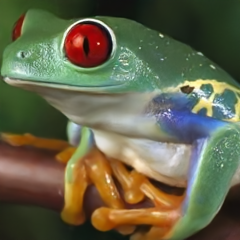} & 
         \includegraphics[trim={0 2cm 0 0},clip, width=0.24\linewidth]{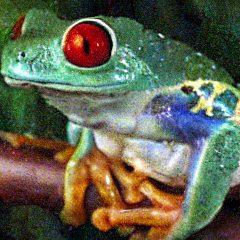} & 
         \includegraphics[trim={0 2cm 0 0},clip, width=0.24\linewidth]{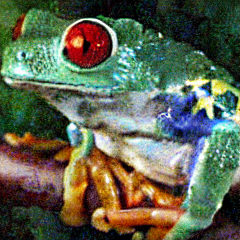} \\
         \midrule
         \rowcolor{lyellow} \multicolumn{4}{c}{Instruction: \emph{``Improve the quality of this image"} -- Real user (Ambiguous) }\\
         \midrule
         \includegraphics[trim={0 2cm 0 0},clip, width=0.24\linewidth]{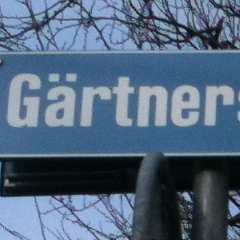} & 
         \includegraphics[trim={0 2cm 0 0},clip, width=0.24\linewidth]{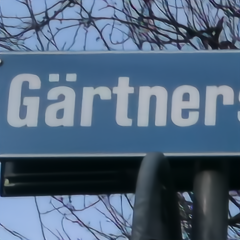} & 
         \includegraphics[trim={0 2cm 0 0},clip, width=0.24\linewidth]{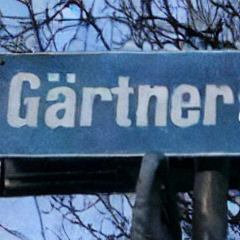} & 
         \includegraphics[trim={0 2cm 0 0},clip, width=0.24\linewidth]{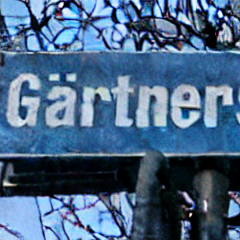} \\
         \midrule
         \rowcolor{lyellow} \multicolumn{4}{c}{Instruction: \emph{``restore this photo, add details"} -- Real user (Precise) }\\
         \midrule
         \includegraphics[trim={0 2cm 0 0},clip, width=0.24\linewidth]{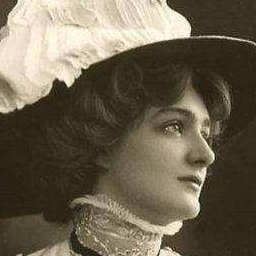} & 
         \includegraphics[trim={0 2cm 0 0},clip, width=0.24\linewidth]{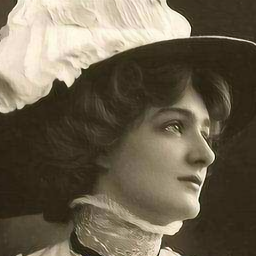} & 
         \includegraphics[trim={0 2cm 0 0},clip, width=0.24\linewidth]{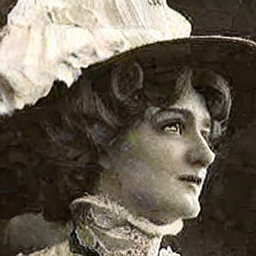} & 
         \includegraphics[trim={0 2cm 0 0},clip, width=0.24\linewidth]{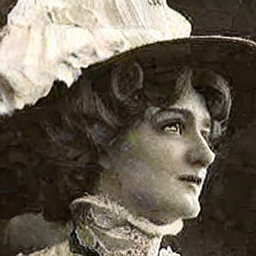} \\
         \midrule
         \rowcolor{lyellow} \multicolumn{4}{c}{Instruction: \emph{``Enhance this photo like a photographer"} -- Basic \& Precise }\\
         \midrule
         \includegraphics[trim={0 2cm 0 0},clip, width=0.24\linewidth]{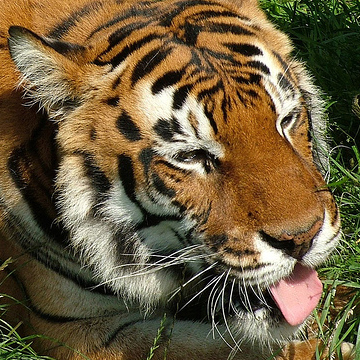} &  
         \includegraphics[trim={0 2cm 0 0},clip, width=0.24\linewidth]{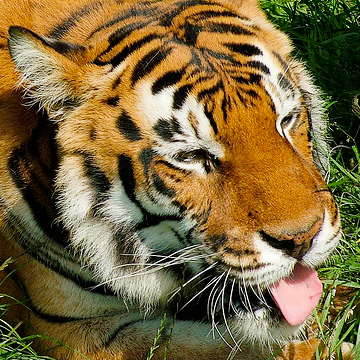} &  
         \includegraphics[trim={0 2cm 0 0},clip, width=0.24\linewidth]{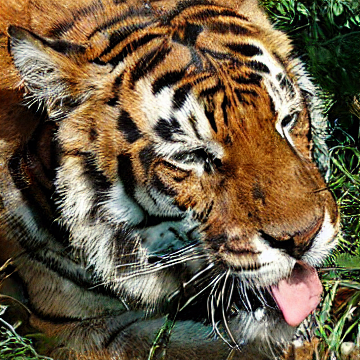} &  
         \includegraphics[trim={0 2cm 0 0},clip, width=0.24\linewidth]{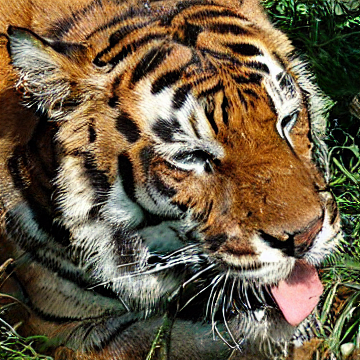} \\
         Input  & \we (ours) & \inst ~~$S_I\!=\!5$ & \inst ~~$S_I\!=\!7$ \\
    \end{tabular}
    \caption{\textbf{ Comparison with \inst~\cite{brooks2023instructpix2pix} for instruction-based restoration using the prompt}. Real-world samples from the \emph{RealSRSet}~\cite{wang2018esrgan, liang2021swinir}. We use our \sevend variant. We run \inst~\cite{brooks2023instructpix2pix} using two configurations where we vary the weight of the image component hoping to improve fidelity: $S_I\!=\!5$ and $S_I\!=\!7$ (also known as Image CFG), this parameters helps to enforce fidelity and reduce hallucinations.}
    \label{fig:comp-instpix2}
\end{figure*}

We present diverse qualitative samples in Figures~\ref{fig:noise-quali},~\ref{fig:rain-quali}. Our method produces high-quality results given images with any of the studied degradations. In most cases the results are better than the reference all-in-one model AirNet~\cite{Li_2022_CVPR}, and the recent SOTA PromptIR~\cite{potlapalli2023promptir}. Also we compare with \instp~\cite{brooks2023instructpix2pix} (diffusion-based) in Figure~\ref{fig:comp-instpix2} using real-world cases. In Figure~\ref{fig:real-haze-q}, we test our method on real-world samples for image dehazing.

\subsection{Efficiency Analysis}

We can \emph{process FHD images under 1s} on consumer-grade GPUs (12-24Gb). We are also notably faster and more efficient than the SOTA method PromptIR~\cite{potlapalli2023promptir} with ~2x less parameters (16M vs. 35M), and ~1.6x less operations.

\begin{table}[!h]
    \vspace{-6mm}
    \centering
    \caption{Inference cost comparison. Some numbers are from~\cite{chen2022simple}.
    }
    \label{tab:infer}
    \resizebox{0.65\linewidth}{!}{
    \begin{tabular}{l|cccccccc}
    % \hline\noalign{\smallskip}
    \toprule
    Method & MPRNet & MIRNet & Restormer & PromptIR & NAFNet & \textbf{InstructIR} \\
    %& \cite{Zamir_2021_CVPR_mprnet} & \cite{zamir2020mirnet} & \cite{wang2021uformer} & \cite{tu2022maxim} & \cite{restormer} & \cite{potlapalli2023promptir} & \cite{chen2022simple} & \textbf{ours} \\
    \midrule
    MACs(G) & 588 & 786 & 140 & 160 & 65 & 100 \\
    \bottomrule
    \end{tabular}
    }
    \vspace{-8mm}
\end{table}%

%%%%%%%%%%%%%%%%%%%%%%%%%%%%%%%%%%%%%%%%%%%%%%%%%%%%%%
%% ORGINAL FIGURES (w ZOOM-IN)

\vspace{-10mm}

%%%%%%%%%%%%%%%%%%%%%%%%%%%%%%%%%%%%%%%%%%%%%%%%%%%%%%

%%%%%%%%%%%%%%%%%%%%%%%%%%%%%%%%%%%%%%%%%%%%%%%%%%%%%%
%%% REAL WORLD DEHAZING

%%%%%%%%%%%%%%%%%%%%%%%%%%%%%%%%%%%%%%%%%%%%%%%%%%%%%%
%% COMP INSTRUCTS

%%%%%%%%%%%%%%%%%%%%%%%%%%%%%%%%%%%%%%%%%%%%%%%%%%


\newpage

\bibliographystyle{splncs04}
\bibliography{main}

\end{document}

%% file: preamble.tex
\usepackage{graphicx}
\usepackage{comment}
\usepackage{booktabs}
\usepackage{tabularx}
\usepackage{multirow}
\usepackage{xcolor,soul}
\usepackage{subcaption}
\usepackage{caption}
\usepackage{makecell}
\usepackage{adjustbox}

\usepackage{pifont}% http://ctan.org/pkg/pifont
\usepackage{xspace}
\def\we{\emph{InstructIR}\xspace}
\def\instp{InstructPix2Pix\xspace}
\def\sota{\emph{state-of-the-art}\xspace}

\def\threed{\textcolor{lpurple}{\bf 3D}\xspace}
\def\fived{\textcolor{purple}{\bf 5D}\xspace}
\def\sixd{\textbf{\textcolor{sblue}{6D}}\xspace}
\def\sevend{\textbf{7D}\xspace}
\def\redast{\textcolor{red}{*}\xspace}

\definecolor{Gray}{gray}{0.9}
\definecolor{lgray}{gray}{0.95}
\definecolor{LightCyan}{rgb}{0.92,0.92,1}
\definecolor{lyellow}{rgb}{1,1,0.92}
\definecolor{lgreen}{rgb}{0.92,1,0.95}
\definecolor{llblue}{rgb}{0.92,0.93,0.95}
\definecolor{lred}{rgb}{1,0.85, 0.85}
\definecolor{tabhighlight}{HTML}{e5e5e5}